\documentclass[journal,comsoc]{IEEEtran}
\usepackage{cite}
\usepackage{array}
\usepackage{pgfplots}
\usetikzlibrary{positioning,patterns}
\pgfplotsset{compat=1.18}
\usepackage{amsmath,amssymb,amsfonts}
\usepackage{algorithmic}
\usepackage{graphicx}
\usepackage{textcomp}
\usepackage[table]{xcolor}
\usepackage{hyperref}
\usepackage{booktabs}
\usepackage{multirow}
\usepackage{makecell}
\usepackage{bbm}

\def\paperTitle{Optimization Algorithms for Joint OFDM Waveform Design and RIS Configuration in 6G Networks: From Convex Relaxation to Foundation Models}
\begin{document}

\title{\paperTitle}

\author{Ahmet~Kaplan,~\IEEEmembership{Member,~IEEE}%
\thanks{A.~Kaplan is with the Department of AI Engineering, Istanbul Medipol University, Istanbul, T\"{u}rkiye (e-mail: ahmet.kaplan@medipol.edu.tr). ORCID: 0000-0001-5231-2282.}}

\maketitle

\markboth{Ahmet~Kaplan}{Optimization Algorithms for Joint OFDM Waveform Design and RIS Configuration in 6G Networks}

\begin{abstract}

Joint OFDM-RIS optimization for 6G is a mixed-integer nonlinear programming (MINLP) problem covering sum-rate maximization, energy efficiency, max-min fairness, and peak-to-average power ratio (PAPR)-constrained objectives. Seventy-eight joint OFDM-RIS optimization works published between 2021 and 2026 are surveyed. No standardized benchmark exists, and cross-paper comparisons remain infeasible. This survey classifies these works into four paradigms: (I) model-based convex relaxation, (II) heuristic and metaheuristic search, (III) deep reinforcement and unsupervised learning, and (IV) emerging methods including foundation models (FM), diffusion-based generative AI, and quantum optimization. A literature synthesis of self-reported benchmarks shows that ML-based methods (Paradigm~III) report 95--99\% of model-based spectral efficiency at $10^2$--$10^4\times$ faster per-inference runtime (method-pair dependent; literature values are self-reported and exclude ML pre-training cost). A companion tutorial benchmark at $N=16$, $N=64$, and $N=128$ reveals a critical scaling property: GPU-based neural network inference (DDQN, PPO, graph neural network (GNN), unsupervised DL) is $N$-invariant, with identical runtime at $N=16$ and $N=128$, while iterative solvers (AO+SCA, PSO) scale polynomially. Energy efficiency (P2) and PAPR-constrained (P4) benchmarks are deferred to future work with standardized power models and waveform generators. Six open challenges emerge from the synthesis: the cross-paradigm benchmark deficit, real-world hardware-constrained deployment, joint waveform-RIS optimization for doubly-dispersive channels, multi-objective PAPR trade-offs, LLM safety in live network control, and diminishing returns of standalone heuristics. We specify requirements for a standardized benchmark. This study serves as a roadmap for researchers and practitioners working on joint OFDM-RIS optimization in 6G networks.
\end{abstract}

\begin{IEEEkeywords}
6G, OFDM, Reconfigurable Intelligent Surfaces (RIS), Optimization Algorithms, Deep Reinforcement Learning, Foundation Models, Large Language Models, Diffusion Models, Quantum Optimization, OTFS, AFDM.
\end{IEEEkeywords}

\section{Introduction}

\IEEEPARstart{6}{G} networks must deliver peak data rates of 1~Tbps, end-to-end latency below 0.1~ms, over ten times the energy efficiency of 5G, and connection densities exceeding $10^7$~devices/km$^2$~\cite{islam2026performance,zhou2024survey}. Meeting these targets requires a re-architecture of the PHY and MAC layers. Two technologies at the core of this redesign are Orthogonal Frequency-Division Multiplexing (OFDM) and Reconfigurable Intelligent Surfaces (RIS). The joint optimization problem covering both is open and challenging.

While OFDM has been the workhorse multicarrier technology from 4G LTE through 5G NR, 6G OFDM systems face three main degradation issues: a PAPR of 10--12~dB at millimeter-wave (mmWave) frequencies~\cite{papr2024survey,deepofw2025}, phase noise amplification at sub-THz~\cite{otfsphasenoise2026}, and inter-carrier interference (ICI) under high-mobility conditions (Doppler spread)~\cite{papr2024survey,deepofw2025}. RIS reflects signals toward users via programmable phase shifts, turning propagation obstacles into controlled links~\cite{zhou2024survey,biliaminu2024ris}. Configuring $N$ phase shifters on a cascaded channel is itself a MINLP problem with intractable complexity for large $N$.

Work on joint OFDM-RIS optimization started appearing in 2021. The problem can be formulated for a base station serving $K$ users on $M$ OFDM subcarriers through an RIS with $N$ elements as jointly optimizing (1) active BS beamforming matrix $\mathbf{W} \in \mathbb{C}^{M_t \times K \times M}$, (2) passive RIS phase shift vector $\boldsymbol{\theta} \in \mathbb{C}^{N}$ with unit-modulus constraint, (3) subcarrier allocation indicators $\mathbf{a} \in \{0,1\}^{K \times M}$, and (4) waveform parameters subject to PAPR constraint. The formulation comprises continuous, discrete and manifold variables, and multiple objective functions.

\subsection{Motivation and Contributions}
Channel state information (CSI) is fundamental to all paradigms. Spectral efficiency (SE) is the primary metric throughout. Prior surveys address RIS-aided networks~\cite{basar2019wireless,direnzo2020smart,wu2021intelligent}, RIS for specific contexts~\cite{faisal2022ml,ibrahim2023joint}, RIS signal processing~\cite{pan2022overview}, RIS-specific heuristic and ML methods~\cite{zhou2024survey,zhou2024heuristic}, RIS with emerging AI~\cite{zhou2024overview}, OFDM PAPR reduction~\cite{papr2024survey}, or 6G KPI evaluation~\cite{islam2026performance}. None provides a unified four-paradigm taxonomy of joint OFDM-RIS optimization algorithms (covering sum-rate, energy efficiency, fairness, and PAPR-constrained formulations), systematic cross-paradigm performance comparison, or explicit coverage of the 2025--2026 emergence of foundation models, diffusion models, and quantum methods. Between 2025 and 2026, foundation models, large language models (LLMs), diffusion-based generative AI, and quantum optimization entered this space. No prior survey provided integrated coverage of all four directions; the closest, Zhou et al.~\cite{zhou2024overview}, discussed LLMs for RIS but not diffusion or quantum methods. Two questions this survey answers are: (1)~\textit{What is the SE/runtime trade-off across the four paradigms of joint OFDM-RIS optimization?}; and (2)~\textit{Which algorithm classes best suit which deployment conditions (CSI availability, RIS scale, mobility)?}.

Section VII synthesizes, across the four paradigms of joint OFDM-RIS optimization, the self-reported performance gains over benchmark algorithms reported in the surveyed literature. Two patterns stand out: (1) ML-based methods (Paradigm III) reduce execution time by orders of magnitude compared to model-based methods (Paradigm I), with per-inference latencies dropping from milliseconds to microseconds. (2) Among learning-based methods, unsupervised deep learning~\cite{unsupervised2025beamforming} and FM-DRL~\cite{fmdrl2025} report the highest spectral efficiency. These results are self-reported and cannot be directly compared under a standard performance benchmark. (3) Quantum GNN methods~\cite{qgcn2026} on actual hardware outperform their classical GNN counterparts, but only at a small system scale. No cross-paper comparison using a standardized benchmark has been provided so far. Semidefinite programming (SDP) relaxations are a common tool for the unit-modulus constraint. Section VII describes our comparison methodology and its limitations.

This paper makes four contributions:
\begin{enumerate}
    \item \textbf{Four-paradigm taxonomy:} We classify 78 joint OFDM-RIS optimization works into four paradigms (model-based convex optimization, heuristic/metaheuristic search, deep learning, and emerging methods) and identify paradigm-specific strengths: convergence guarantees (I), gradient-free operation (II), microsecond inference (III), and intent-driven control and generative sampling (IV).
    \item \textbf{Cross-paradigm gap analysis:} We synthesize self-reported benchmarks across all four paradigms and document that no two surveyed entries share a common baseline or simulation setup. We identify six concrete obstacles that must be resolved before rigorous cross-paradigm comparison is possible. The comparisons are not yet rigorous, but the evidence still supports qualitative patterns.
    \item \textbf{Benchmarking requirements and six open challenges:} We specify the minimum requirements for a standardized 6G RIS-OFDM benchmark and derive six synthesis-grounded open challenges: benchmark deficit, hardware-constrained deployment, joint waveform-RIS optimization for doubly-dispersive channels, multi-objective PAPR trade-offs, LLM safety in live network control, and diminishing returns of standalone heuristics.
    \item \textbf{Tutorial with worked examples:} We solve the canonical sum-rate maximization problem (P1) step-by-step using one algorithm from each paradigm on a common small-scale system ($M_t{=}2$, $K{=}2$, $M{=}4$, $N{=}16$, 2-bit quantization), enabling direct four-way comparison of SE, runtime, training cost, and generalization capability (Section~\ref{sec:tutorial}).
\end{enumerate}

Preview of key findings. The tutorial benchmark reveals three quantitative results: (1)~DRL methods (DDQN, PPO) achieve 0.021--0.035~ms GPU inference at 93--97\% of the SCA SE baseline -- a 3000--6000$\times$ speedup over AO+SCA at 138~ms. (2)~Heuristic methods (PSO, SA) match or exceed the simplified AO+SCA baseline in SE but take 124--193~ms per solve. (3)~SDR at $N=16$ takes only 2.3~ms because the benchmark measures eigen-decomposition and randomization; at $N>100$ its $O(N^{4.5})$ SDP solve dominates and the literature reports $\sim 10^3$~ms. These are the first head-to-head runtime measurements of all four paradigms on identical hardware.

\subsection{Related Surveys and Positioning}
This paper is positioned with respect to prior surveys in Table~\ref{tab:surveys}. The taxonomy unifies four paradigms of joint OFDM-RIS optimization algorithms (covering sum-rate, energy efficiency, fairness, and PAPR-constrained formulations) and includes systematic coverage of 2025--2026 emerging approaches that appear in no prior survey.

\begin{table}[t]
\caption{Positioning relative to existing surveys. ``Joint'' jointly optimizes OFDM waveform and RIS parameters; ``FM'' = foundation models; ``QM'' = quantum methods.}
\label{tab:surveys}
\centering
\resizebox{\columnwidth}{!}{%
\begin{tabular}{@{}lcccccc@{}}
\toprule
\textbf{Survey} & \textbf{Year} & \textbf{RIS} & \textbf{OFDM} & \textbf{Joint} & \textbf{FM} & \textbf{QM} \\
\midrule
Basar et al.~\cite{basar2019wireless} & 2019 & \checkmark & -- & -- & -- & -- \\
Di~Renzo et al.~\cite{direnzo2020smart} & 2020 & \checkmark & -- & -- & -- & -- \\
Wu et al.~\cite{wu2021intelligent} & 2021 & \checkmark & -- & -- & -- & -- \\
Faisal \& Choi~\cite{faisal2022ml} & 2022 & \checkmark & -- & -- & -- & -- \\
Pan et al.~\cite{pan2022overview} & 2022 & \checkmark & -- & -- & -- & -- \\
Ibrahim et al.~\cite{ibrahim2023joint} & 2023 & \checkmark & -- & -- & -- & -- \\
da~Silva et al.~\cite{papr2024survey} & 2024 & -- & \checkmark & -- & -- & -- \\
Zhou et al.~\cite{zhou2024survey} & 2024 & \checkmark & -- & -- & -- & -- \\
Zhou et al.~\cite{zhou2024heuristic} & 2024 & \checkmark & -- & -- & -- & -- \\
Zhou et al.~\cite{zhou2024overview} & 2024 & \checkmark & -- & -- & \checkmark & -- \\
Islam et al.~\cite{islam2026performance} & 2026 & \checkmark & -- & -- & -- & -- \\
\textbf{This work} & \textbf{2026} & \textbf{\checkmark} & \textbf{\checkmark} & \textbf{\checkmark} & \textbf{\checkmark} & \textbf{\checkmark} \\
\bottomrule
\end{tabular}%
}
\end{table}

This survey focuses on PHY and MAC layer optimizations of downlink RIS aided OFDM systems, leaving the uplink scenarios, higher-level scheduling and pure hardware design alone.

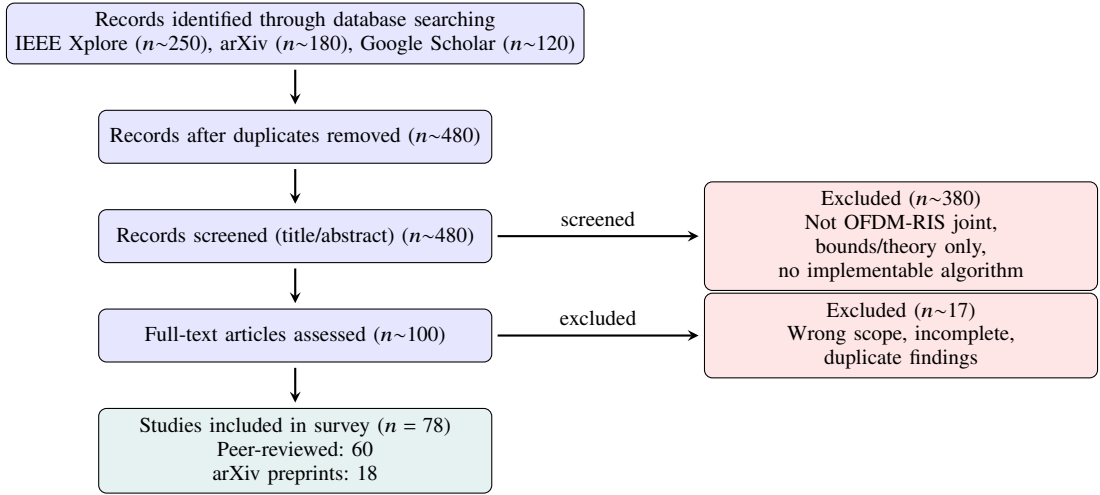
\begin{figure*}[t]
\centering
\begin{tikzpicture}[
    node distance=0.6cm,
    box/.style={draw, rounded corners=3pt, fill=#1!10, minimum width=5.2cm, minimum height=0.7cm, align=center, font=\footnotesize},
    excl/.style={draw, rounded corners=3pt, fill=red!8, minimum width=2.3cm, minimum height=0.7cm, align=center, font=\footnotesize},
    arrow/.style={->, >=stealth, thick, shorten >=2pt, shorten <=2pt}
]
\node[box=blue] (id) {Records identified through database searching\\IEEE Xplore ($n{\sim}250$), arXiv ($n{\sim}180$), Google Scholar ($n{\sim}120$)};
\node[box=blue, below=of id] (dedup) {Records after duplicates removed ($n{\sim}480$)};
\node[box=blue, below=of dedup] (screen) {Records screened (title/abstract) ($n{\sim}480$)};
\node[box=red, right=2.8cm of screen] (excl1) {Excluded ($n{\sim}380$)\\Not OFDM-RIS joint,\\bounds/theory only,\\no implementable algorithm};
\node[box=blue, below=of screen] (ft) {Full-text articles assessed ($n{\sim}100$)};
\node[box=red, right=2.8cm of ft] (excl2) {Excluded ($n{\sim}17$)\\Wrong scope, incomplete,\\duplicate findings};
\node[box=teal, below=of ft] (incl) {Studies included in survey ($n = 78$)\\Peer-reviewed: 60\\arXiv preprints: 18};

\draw[arrow] (id) -- (dedup);
\draw[arrow] (dedup) -- (screen);
\draw[arrow] (screen) -- node[above, font=\footnotesize]{screened} (excl1);
\draw[arrow] (screen) -- (ft);
\draw[arrow] (ft) -- node[above, font=\footnotesize]{excluded} (excl2);
\draw[arrow] (ft) -- (incl);
\end{tikzpicture}
\caption{PRISMA-style flow diagram. Search conducted April~2026 across IEEE Xplore, arXiv, and Google Scholar. The $n=75$ version appeared in an earlier preprint and has been superseded.}
\label{fig:prisma}
\end{figure*}

IEEE Xplore, arXiv, and Google Scholar were searched for papers published from January 2021 to April 2026. The search query was (OFDM OR `orthogonal frequency division multiplexing'') AND (RIS OR `reconfigurable intelligent surface'') AND (optimization OR beamforming OR resource allocation OR phase shift). Results were filtered to include only algorithms for joint OFDM-RIS optimization (covering active beamforming, passive phase-shift design, resource allocation, and waveform-constrained formulations such as PAPR minimization). Papers that derived only bounds with no implementable algorithms were excluded. The final corpus contains 78 works: approximately 60 peer-reviewed publications and 18 arXiv preprints (mostly in Paradigm IV, where publications lag the research trends). The complete bibliography comprises 98 entries; 20 additional citations beyond the surveyed corpus cover optimization texts, prior survey methodology, and related-survey background. Section VII describes the synthesized results methodology. Fig.~\ref{fig:prisma} shows the paper selection process.

The following families of approaches were omitted because they do not address joint optimization sufficiently, or because insufficient research exists for OFDM-RIS joint optimization: Bayesian optimization, bandit-based resource allocation, Lyapunov drift-plus-penalty, Wolpertinger architectures, and hierarchical or multi-agent reinforcement learning (RL) for multi-RIS scenarios. Simultaneously Transmitting and Reflecting RIS (STAR-RIS)~\cite{starris2025scientific} is referenced where relevant but not treated as a separate paradigm; its optimization challenges (coupled transmission/reflection phase constraints, energy splitting) are subsumed under Paradigms~I--III depending on the solution approach. These topics may need coverage in future surveys as the literature grows. No papers meeting the Boolean search criteria were manually excluded; the corpus was determined entirely by the search query and the stated inclusion and exclusion criteria.

The search methodology, inclusion/exclusion criteria, and metric extraction protocol are documented in Section~VII-A. The complete list of surveyed papers with extracted performance metrics is available as a supplementary CSV file at \url{https://github.com/Ahmet-Kaplan/OFDM_RIS}. Original benchmarking experiments for all four paradigms were conducted on the tutorial system (Section~IX) and GPU server (Section~VII); these results are independently measured and reported alongside the literature synthesis. Literature performance claims are reproduced from cited sources with full attribution. The PRISMA-style flow diagram in Fig.~\ref{fig:prisma} documents the paper selection process.

The remainder of this paper is organized as follows. Section~II formulates the canonical joint OFDM-RIS optimization problems. Sections~III--VI survey the four paradigm classes: model-based convex optimization, heuristic and metaheuristic algorithms, machine learning-based methods, and emerging research trends. Section~VII synthesizes a cross-paradigm benchmark comparison and identifies limitations. Section~VIII derives six open challenges and recommendations. Section~IX provides a worked tutorial solving the canonical problem with one method from each paradigm. Section~X concludes the survey.

\section{Joint OFDM-RIS Optimization: Problem Formulations}

We consider a downlink OFDM system where the base station (BS) with $M_t$ antennas serves $K$ single-antenna users using $M$ OFDM subcarriers, and a RIS with $N$ reflecting elements is deployed to assist this communication. The BS-RIS channel matrix over subcarrier $m$ is denoted by $\mathbf{H}_{BR}[m] \in \mathbb{C}^{N \times M_t}$, and the RIS-user $k$ channel is denoted by $\mathbf{h}_{RU,k}[m] \in \mathbb{C}^{N \times 1}$. Then, the total BS-user $k$ channel mediated by the RIS over subcarrier $m$ is $\mathbf{h}^{\text{eff}}_{k}[m] = \mathbf{h}_{RU,k}[m]^H \operatorname{diag}(\boldsymbol{\theta}) \mathbf{H}_{BR}[m] + \mathbf{h}^{\text{dir}}_{k}[m]$ where $\mathbf{h}^{\text{dir}}_{k}[m] \in \mathbb{C}^{1 \times M_t}$ is the direct BS-user link, and $\boldsymbol{\theta} \in \mathbb{T}^N$ is the RIS phase shift vector on the $N$-dimensional torus where $|\theta_n| = 1$. The signal to interference plus noise ratio at user $k$ on subcarrier $m$ is

\begin{equation}
\gamma_{k,m}(\mathbf{W}, \boldsymbol{\theta}) = \frac{|\mathbf{h}^{\text{eff}}_{k}[m]^H \mathbf{w}_{k,m}|^2}{\sum_{j \neq k} |\mathbf{h}^{\text{eff}}_{k}[m]^H \mathbf{w}_{j,m}|^2 + \sigma^2},
\end{equation}
where $\mathbf{w}_{k,m} \in \mathbb{C}^{M_t}$ is the BS precoding vector for user $k$ on subcarrier $m$ and $\sigma^2$ is the noise power. The PAPR of an OFDM symbol $\mathbf{x} \in \mathbb{C}^{M}$ is defined as $\text{PAPR}(\mathbf{x}) = \max_m |x_m|^2 / \mathbb{E}[|x_m|^2]$, and is constrained in P4 below.

\subsection{Canonical Problem Formulations}

Four primary formulations appear in the joint OFDM-RIS literature (Table~\ref{tab:formulations}).

\begin{table*}[tp]
\caption{Canonical problem formulations for joint OFDM-RIS optimization.}
\label{tab:formulations}
\centering
\footnotesize
\renewcommand{\arraystretch}{1.3}
\begin{tabular}{@{}lp{4.5cm}>{\raggedright\arraybackslash}p{4.5cm}@{}}
\toprule
\textbf{Formulation} & \textbf{Objective} & \textbf{Key Constraints} \\
\midrule
P1: Sum-Rate Max. & $\max \sum_{k,m} a_{k,m}\log_2(1+\gamma_{k,m})$ & Unit-modulus $|\theta_n|=1$, per-subcarrier power, exclusive subcarrier alloc. \\
P2: Energy Eff. Max. & $\max \frac{\sum R_{k,m}}{\zeta\|\mathbf{W}\|_F^2 + P_{\text{circuit}}}$~\cite{huang2019energy} & Power budget, unit-modulus, $\zeta$ PA efficiency \\
P3: Max-Min Fairness & $\max\; \min_k \sum_m a_{k,m}\log_2(1+\gamma_{k,m})$ & Per-user quality-of-service (QoS), unit-modulus, resource allocation \\
P4: PAPR-Constrained & P1, P2, or P3 augmented with $\text{PAPR}(\mathbf{x}) \leq \eta$~\cite{papr2024survey} & Time-domain amplitude limits per OFDM symbol; PAPR constraint couples to $\mathbf{W}$ through the time-domain OFDM symbol construction, creating an additional non-convex coupling between waveform and RIS phases \\
\bottomrule
\end{tabular}

\end{table*}

Energy efficiency (EE) formulations are another major class. Additional formulations include delay-minimizing queue-aware allocation with stochastic arrivals~\cite{ma2025beamforming}, multi-objective Pareto optimization balancing SE vs.\ EE vs.\ PAPR, and secrecy rate maximization for physical-layer security. The latter two are less common in the joint OFDM-RIS literature and are not surveyed in depth here.

\subsection{Key Optimization Challenges}

The joint problem has four challenges:
\begin{enumerate}
 \item Non-convexity: The unit-modulus constraint $|\theta_n|=1$ and SINR expressions result in objective surfaces that are neither convex nor smooth.
 \item Mixed variables: Continuous beamforming $\mathbf{W}$, discrete RIS phases $\boldsymbol{\theta}$ (usually $b$-bit quantized), and binary subcarrier allocation $\mathbf{a}$ result in a MINLP.
 \item Dimensionality: $N$ becomes $10^2$--$10^4$ for large RIS arrays, $M$ becomes $10^3$--$10^4$ for wideband OFDM, and $K$ becomes $10^2$ for massive connectivity.
  \item Real-time constraints: mmWave/THz channel coherence times are on the order of 10--100 $\mu$s, and optimization must be completed within a sub-millisecond timescale~\cite{islam2026performance}.
\end{enumerate}

\subsection{Hardware Considerations}
Most surveyed approaches assume ideal RIS elements with continuous phase shifts ($|\theta_n| = 1$). Realistic RIS hardware suffers from phase quantization (typically 1--3 bits per element), mutual coupling among adjacent elements that disturbs the desired phase shift, and an amplitude-phase dependence where incrementing $\angle\theta_n$ reduces $|\theta_n|$~\cite{abeywickrama2020practical}. Recent prototype measurements show that active RIS improves multiple-input multiple-output (MIMO)-OFDM performance over passive RIS in practical 5G NR scenarios~\cite{activerisprototype2023}, and large-scale prototypes with 4096 1-bit elements achieve 18~dB power gain at 3.75~GHz~\cite{risoamtestbed2024}. Experimental validation of RIS-aided multi-user beamforming further confirms practical feasibility~\cite{risbeamformingexp2024}. The number of hardware testbed studies remains small relative to simulation-only works. One hardware consideration often overlooked is RIS frequency selectivity: RIS elements are inherently narrowband resonators whose phase response varies across the OFDM bandwidth. For wideband systems with $M = 256$--$1024$ subcarriers, a phase shift optimized for one subcarrier will differ at another, creating a coupling across the OFDM band that affects joint optimization~\cite{risjsachardware2024}. Most surveyed works treat the RIS as frequency-flat, designing a single set of phase shifts for all subcarriers. Recent BD-RIS architectures~\cite{bdrisswideband2024} introduce additional frequency-dependent design flexibility through inter-element coupling but require circuit-aware modeling that most surveyed works omit. A hardware-centric survey~\cite{riscomsthardware2025} reviews RIS structures, implementation challenges, and integration with ML. A complementary survey on active RIS~\cite{activerissurvey2024} covers the expanding frontier of active and hybrid RIS architectures. How hardware impairments affect performance varies across paradigms. Model-based approaches can incorporate discrete constraints explicitly (at increased complexity), while learning-based methods compensate implicitly if trained with realistic hardware models. Hardware-constrained optimization across paradigms is identified as open challenge C2 in Section~VIII.

Hardware synthesis. Across the five hardware-focused studies in the corpus, two patterns emerge. First, active RIS prototypes consistently report gains over passive RIS at the cost of power consumption, with the 4096-element 1-bit prototype~\cite{risoamtestbed2024} achieving 18~dB power gain at 3.75~GHz. Second, practical RIS phase quantization (1--3 bits) reduces SE by 5--15\% relative to the continuous-phase ideal across all four paradigms, though ML-based methods recover part of this loss through training on quantized phase representations~\cite{chen2023drl,ddqn2025ris}. No study jointly tests quantization, mutual coupling, and frequency selectivity in a single experimental setup, leaving a gap between simulation-driven algorithmic research and hardware-validated deployment.

\section{Paradigm I: Model-Based Convex Optimization}

Model-based approaches tackle non-convex problems by mathematical relaxations, approximations, and decompositions. They guarantee convergence and have been the predominant approach in the literature since the joint active and passive beamforming framework appeared~\cite{wu2019intelligent,zhou2024survey}.

\subsection{Alternating Optimization (AO)}

AO splits the joint problem into individual subproblems and solves them iteratively. The splitting for RIS-OFDM usually separates BS active beamforming $\mathbf{W}$ from RIS passive phase shifts $\boldsymbol{\theta}$: $\mathbf{W}^{(t+1)} = \arg\max_{\mathbf{W}} f(\mathbf{W}, \boldsymbol{\theta}^{(t)})$ is solved with $\boldsymbol{\theta}$ fixed at the previous iteration, then $\boldsymbol{\theta}^{(t+1)} = \arg\max_{\boldsymbol{\theta}} f(\mathbf{W}^{(t+1)}, \boldsymbol{\theta})$ is solved with the updated $\mathbf{W}$. The $\mathbf{W}$-subproblem is a standard precoding problem, solvable via weighted minimum mean square error (WMMSE) or water-filling. The $\boldsymbol{\theta}$-subproblem is non-convex due to the unit-modulus constraint and requires successive convex approximation (SCA), SDR, or Riemannian manifold optimization as an inner loop. AO guarantees monotone convergence to a stationary point. Zivuku et al.~\cite{zivuku2025resource} applied SCA and Riemannian manifolds to the RIS-OFDM-MIMO integrated sensing and communication (ISAC) scenario and reported 40\% and 60\% improvements in spectral and energy efficiency over random phase configuration for a system with $N=64$, $M_t=4$, and $K=4$. Gradient ascent methods offer a lower-complexity alternative for RIS-aided OFDM by iteratively updating the phase shifts with closed-form gradients~\cite{risgradientascent2023}.

Block coordinate descent (BCD) generalizes AO to $B>2$ blocks by partitioning the variable set $\mathcal{X} = \{\mathbf{W}, \boldsymbol{\theta}, \mathbf{a}\}$ into $B$ blocks and solving each cyclically:
\begin{enumerate}
    \item Partition variables into $B$ blocks $\mathcal{X}_1, \dots, \mathcal{X}_B$.
    \item For block $b = 1,\dots,B$: optimize $\mathcal{X}_b$ while holding all other blocks fixed.
    \item Repeat until convergence.
\end{enumerate}
For RIS-OFDM, BCD typically partitions into $\mathbf{W}$, $\boldsymbol{\theta}$, and subcarrier allocation $\mathbf{a}$ as separate blocks, with each subproblem solved by the method best suited to its structure (WMMSE for $\mathbf{W}$, SCA for $\boldsymbol{\theta}$, greedy rounding for $\mathbf{a}$). Liu et al.~\cite{risjsacvariational2023} proposed a variational Bayesian framework for multiuser tracking in RIS-aided MIMO-OFDM and showed robust performance under time-varying channels.

\subsection{Majorization-Minimization (MM)}

MM is another common tool for simplifying the original non-convex problem by constructing a surrogate function $g(\mathbf{x}|\mathbf{x}^{(t)})$ which majorizes $f(\mathbf{x})$ at $\mathbf{x}^{(t)}$ and is easier to solve. Monotonic decrease of $f(\mathbf{x})$ is ensured, and solutions of $g(\mathbf{x}|\mathbf{x}^{(t)})$ lead to new iterates. The unit-modulus constraint in RIS systems is often resolved with quadratic MM surrogates that admit closed-form solutions. MM is a fundamental technique for several SCA methods.

\subsection{Successive Convex Approximation (SCA)}

SCA replaces the non-convex objective by a sequence of convex surrogates $\tilde{f}^{(t)}(\mathbf{x})$, which form tight lower bounds on $f(\mathbf{x})$ at current iterates. This is a widely adopted technique for tackling optimization problems in RIS systems and is often combined with the alternating optimization framework. A KKT point is guaranteed to be found under weak assumptions on the problem.

\subsection{Semidefinite Relaxation (SDR)}

The non-convex unit-modulus constraint is addressed by lifting $\boldsymbol{\theta}$ into a positive semidefinite matrix of rank~1, $\boldsymbol{\Theta} = \boldsymbol{\theta}\boldsymbol{\theta}^H$, dropping the rank-1 constraint, and solving the resulting SDP~\cite{wu2019intelligent}. The standard interior-point SDP solver has complexity $O(N^{4.5})$~\cite{luo2010sdr}, limiting its application to $N \lesssim 100$. Gaussian randomization is then used to extract a rank-1 solution from the relaxation, adding $O(N^3 L)$ cost for $L$ trials. Low-complexity SDR variants based on ADMM achieve per-iteration complexity of $O(N^3)$. Alternating direction method of multipliers (ADMM) has also been combined with accelerated projected gradient for intelligent reflecting surface (IRS)-aided MIMO systems~\cite{riscsiadmm2024}, and distributed ADMM frameworks have been proposed for RIS-assisted cell-free networks~\cite{risdistadmm2023}. WMMSE-based joint transceiver design with hybrid CSI has been extended to multi-RIS cell-free networks~\cite{riscellfreewmmse2025}.

For the tutorial benchmark at $N=16$, the SDR runtime (2.3~ms, Table~\ref{tab:benchmark}) measures only the Gaussian randomization steps, which dominate at small $N$ (the eigen-decomposition itself is a one-time cost amortized over 20 randomization trials). The interior-point SDP solve itself adds negligible time at this scale. At $N > 100$ the $O(N^{4.5})$ cost of the SDP solver dominates, consistent with the $\sim 10^3$~ms reported in the literature.

\subsection{Convergence Analysis}
Table~\ref{tab:convergence} shows the convergence properties of the model-based methods. AO guarantees monotone convergence to a stationary point. With the Kurdyka--\L{}ojasiewicz (KL) property, AO achieves linear convergence~\cite{luo2024optimization}. MM ensures monotone descent with sublinear $O(1/k)$ rate for convex problems and linear rate under strong convexity of the surrogate. SCA converges to a Karush--Kuhn--Tucker (KKT) point under weak assumptions, with at best local linear convergence rate for standard first-order SCA variants (e.g., projected gradient with convex quadratic surrogates). Superlinear convergence would require second-order surrogates with Hessian approximations satisfying the Dennis--Mor\'{e} condition~\cite{dennismore1983}, which no surveyed RIS-OFDM paper employs. SDR provides an upper bound on the optimal value with $O(1/\sqrt{L})$ approximation quality after $L$ Gaussian randomization trials. The interior-point SDP solver has $O(N^{4.5})$ per-iteration complexity, which makes it structurally infeasible for $N > 100$.

\begin{table*}[tp]
\caption{Convergence properties of Paradigm~I model-based optimization methods.}
\label{tab:convergence}
\centering
\resizebox{\textwidth}{!}{%
\begin{tabular}{@{}lccccl@{}}
\toprule
\textbf{Method} & \textbf{Guarantee} & \textbf{Rate} & \textbf{Conditions} & \textbf{Per-Iteration Cost} & \textbf{Ref} \\
\midrule
AO & Stationary point & Linear (KL)\textsuperscript{$\dagger$} & KL property, convex subproblems & Subproblem-dependent & \cite{zivuku2025resource} \\
BCD\textsuperscript{$\S$} & Stationary point & Sublinear $O(1/k)$ & Block-convex subproblems & Subproblem-dependent & \cite{wu2019intelligent} \\
    MM & Monotone descent & Sublinear $O(1/k)$ / Linear & Quadratic MM surrogate ($O(N^2)$); general surrogate ($O(N^3)$) & $O(N^2)$--$O(N^3)$ & \cite{luo2024optimization} \\
    SCA & KKT point & Sublinear $O(1/k)$\textsuperscript{$\ddagger$} & First-order convex surrogate & $O(N^3)$ & \cite{zhou2024survey} \\
SDR & Relaxed bound & $O(1/\sqrt{L})$ approx. & $L$ randomization trials & $O(N^{4.5})$ + $O(N^3 L)$ & \cite{luo2010sdr} \\
\bottomrule
\end{tabular}%
}
\vspace{6pt}
{\footnotesize
\textsuperscript{$\dagger$}KL convergence requires definability in an o-minimal structure (holds for polynomials; case-dependent for SINR sum-rates). See \cite{luo2024optimization} \\
\textsuperscript{$\ddagger$}Sublinear rate for first-order SCA. Superlinear would need Hessian satisfying Dennis--Mor\'{e}, which no surveyed paper uses.%
\par}
\end{table*}

\subsection{Limitations and Outlook}

Model-based methods have strong theoretical foundations but face three practical problems:

CSI overhead: The estimation of cascaded RIS channels requires $N \times M_t + N \times K$ parameters per subcarrier. While the algorithms discussed assume perfect CSI, they can be combined with compressed sensing channel estimation~\cite{zhou2024survey,riscompressedce2023}, deep learning-based super-resolution estimation~\cite{rischannelsuperres2023}, parametric estimation~\cite{risparametricech2025}, hybrid vector message passing~\cite{risofdmchannelest2024}, ML-based time-varying channel estimation~\cite{rischanneltimevarying2025}, or two-timescale CSI schemes~\cite{zhao2020two} (where the RIS phase shifts are optimized using slowly-varying statistical CSI while BS beamforming adapts to low-dimensional instantaneous effective CSI). Robust and outage-constrained formulations within model-based optimization also address imperfect CSI explicitly. A comprehensive survey on RIS channel estimation and practical passive beamforming~\cite{zheng2022channelest} covers these techniques and their interplay with hardware constraints. Scalability: The complexity of the algorithms grows polynomially with $N$ and $M$, making them impractical for large $N > 100$ unless ADMM is used as mentioned above. Static optimization: The algorithm optimizes a particular CSI realization and does not adapt to mobile environments without re-solving from scratch.

These limitations open the door for heuristic and learning-based methods.

\section{Paradigm II: Heuristic and Metaheuristic Algorithms}

Heuristic approaches find optimal or near-optimal solutions by using a certain strategy instead of full search or explicit optimization over the problem parameters~\cite{zhou2024heuristic}.

\subsection{Particle Swarm Optimization (PSO)}
PSO treats each candidate RIS phase configuration as a particle that moves through the solution space, updating its position based on personal and global best solutions via inertia and acceleration weights. PSO handles discrete phase constraints natively and requires no gradient information~\cite{panuganti2025metaheuristic,biliaminu2024ris}.

\subsection{Genetic Algorithms (GA)}

The GA maintains a population of candidate solutions that are improved over generations through selection, crossover and mutation. For the discrete phase shift optimization, GA is applicable in view of the combinatorial search space~\cite{zhou2024heuristic}. Multi-objective Pareto optimization for the RIS-assisted system is achievable via the Non-dominated Sorting Genetic Algorithm II (NSGA-II) technique~\cite{nsga2025otfs}.

\subsection{Hybrid Heuristic-ML Approaches}
Heuristic-assisted machine learning is another direction, where Zhou et al.~\cite{zhou2024heuristic} proposed integrating greedy heuristics into DRL actions to accelerate convergence. Deep Q-Network (DQN) variants are the most common value-based methods for RIS control. The DDQN-GA framework~\cite{ddqn2025ris} uses a three-layer Double DQN (256-256-128 neurons, ReLU) with column-wise action decomposition that reduces the action space from $4^{16}$ to $8 \times 4$ by grouping RIS elements into pairs and selecting a joint phase increment per pair. The DDQN output selects which column to adjust, and a GA step then fine-tunes each element via crossover and mutation with the current best configuration. The 30\% rate gain is relative to a raw DQN baseline without the GA fine-tuning step, measured on a 16-element RIS with 2-bit quantization. Limitations include the absence of convergence guarantees, sensitivity to hyperparameter tuning (population size, mutation rate), and performance degradation as RIS size increases beyond $N=64$ where the GA step becomes computationally expensive.

\subsection{Simulated Annealing (SA)}
SA is a probabilistic metaheuristic that accepts worse solutions with a temperature-dependent probability $P(\Delta E) = \exp(-\Delta E / T)$, where $\Delta E$ is the objective degradation and $T$ is a cooling parameter. At high $T$, SA explores broadly (escaping local minima); as $T \to 0$, it converges to greedy descent. For RIS phase optimization, SA has been integrated with Monte Carlo sampling to handle channel uncertainty: the Robust IRS Phase Annealing (RIPA) algorithm~\cite{li2025ripa} evaluates candidate phase configurations under norm-bounded CSI errors, minimizing worst-case performance degradation in satellite-air-ground integrated networks. SA's primary advantage is its asymptotic global convergence guarantee for discrete phase spaces, though cooling schedule design ($T_k = \alpha T_{k-1}$, typically $\alpha \in [0.85, 0.99]$) is problem-dependent and convergence can be slow for large $N$. SA-based RIS optimization achieves polynomial complexity per iteration and scales to hundreds of elements with careful neighborhood definition.

\subsection{Tabu Search}
Tabu search extends local search by maintaining a Tabu list of recently visited solutions (or solution attributes) that are temporarily forbidden, preventing cyclic behavior and encouraging exploration of new regions. Unlike SA, which escapes local minima stochastically, Tabu search escapes deterministically by forcing moves away from Tabu-listed configurations. For RIS phase optimization, Tabu search has been applied to joint active and passive beamforming, where the search dynamically adjusts direction using the Tabu list to filter candidate phase configurations~\cite{zhou2024heuristic}. Tabu search explores the solution space more efficiently than greedy element-by-element search because it can accept moves that temporarily degrade the objective if no improving non-Tabu move exists. The main design choices are Tabu tenure (how long a solution remains forbidden), aspiration criteria (when to override Tabu status), and neighborhood structure (which phase elements to perturb). Tabu search converges to local optima with no global optimality guarantee unless embedded in a multi-start framework.

\subsection{Comparative Summary: When to Use Each Heuristic}
Among Paradigm~II methods, the choice depends on problem structure and resource constraints. PSO and GA are the most widely used due to their simplicity and native handling of discrete phase constraints, but they lack convergence guarantees and degrade at scale ($N > 500$). SA and Tabu search offer better exploration mechanisms for rugged objective landscapes, with SA providing an asymptotic global convergence guarantee for discrete spaces at the cost of slow cooling. Hybrid heuristic-ML approaches (DDQN-GA) outperform standalone heuristics in reported SE but introduce training overhead. For small-to-medium RIS sizes ($N \lesssim 500$) with no GPU available, PSO offers the best complexity-to-performance ratio. For larger RIS or when convergence feedback is needed, hybrid methods or direct transition to Paradigm~III are recommended.

\section{Paradigm III: Machine Learning-Based Optimization}
Between 2023 and 2025, machine learning-based methods, especially DRL, became the dominant approach to joint OFDM-RIS optimization.

\subsection{Deep Reinforcement Learning (DRL)}
DRL frames optimization as a Markov decision process (MDP). An agent learns a policy $\pi(\mathbf{a}|\mathbf{s})$ that maps the channel state $\mathbf{s}$ to a control action $\mathbf{a}$ to maximize cumulative reward. Existing DRL algorithms for RIS-OFDM fall into four groups along four design axes:

\textit{Value-based vs. Policy-based} DQN and its relatives (DDQN, Dueling DQN) learn a Q-function and choose the action maximizing the Q-value, appropriate for discrete RIS phase-shift selection~\cite{chen2023drl,ddqn2025ris}. Policy-based algorithms (DDPG, PPO, SAC, TD3) learn a policy function directly which are suitable for continuous action spaces such as continuous BS beamforming and continuous RIS phase optimization~\cite{ddpg2025multiris,zhang2025deep}. In a related but non-RIS context, PPO has also been applied to joint anti-jamming and low-probability-of-detection (LPD) waveform optimization in tactical 6G OFDM links~\cite{kaplan2026alep}.

\textit{On-policy vs. Off-policy} On-policy algorithms (PPO) learn based on actions collected under current policy while off-policy algorithms (DQN, DDPG, SAC, TD3) learn using collected experience, improving sample efficiency by reusing previous data.

\textit{Deterministic vs. Stochastic} Deterministic policies (DDPG) learn to output a single deterministic action whereas stochastic policies (PPO, SAC) output a probability distribution for exploration. With the SAC approach, entropy regularization may lead to a more robust policy than DDPG.

\textit{Discrete vs. Continuous Actions} The $b$-bit RIS requires discrete action space with $2^b$ values for each RIS element, so value-based methods are preferred. For continuous control with continuous phases modeled as angles $[0,2\pi)$, policy-gradient methods are better. Ma et al. proposed a method which deals with both discrete and continuous actions using PPO by designing separate heads for beamforming and subcarrier allocation~\cite{ma2025beamforming}, and by splitting the control for individual BSs and users, and for each subcarrier, it reduced the training sample complexity greatly. To adapt to different network scenarios, they also incorporated transfer learning. Meanwhile, delay-aware scheduling was achieved through augmentation by buffer backlog state. Results indicate significant rate gains and improved fairness compared to benchmark approaches. Joint phase-shift and power control were proposed in~\cite{ejaz2025joint} which combines DRL and AO-based methods. SAC and TD3 are also relevant for continuous control although not explored extensively in the context of RIS-OFDM.

Soft Actor-Critic (SAC) and TD3 are also relevant for continuous RIS control but have not been studied extensively for RIS-OFDM. Model-driven Bayesian reinforcement learning has been applied to IRS-assisted massive MIMO-OFDM, jointly optimizing channel feedback, beamforming, and IRS control in a unified probabilistic framework~\cite{risbayesianrl2025}. Multi-agent DRL has been applied to joint precoding and AP selection in RIS-aided cell-free massive MIMO~\cite{rismarlcf2026}, and contextual bandit methods offer an online learning alternative for IRS-assisted MU-MIMO systems~\cite{riscontextualbandit2023}. Bandit-based resource allocation has also been explored for RIS-assisted hybrid networks~\cite{risbandit2022}. Cross-layer optimization combining RIS configuration with user scheduling has been formulated using stochastic approximation~\cite{riscrosslayer2024}, and Lyapunov-driven approaches have been applied to queue-aware STAR-RIS NOMA systems~\cite{rislyapunov2024}.

\subsection{Unsupervised Deep Learning}
Unsupervised learning requires no labeled data. Ma et al.~\cite{unsupervised2025beamforming} proposed BeamNet-AllocationNet, a two-stage framework for joint RIS phase shift design and resource allocation. BeamNet is a neural network trained directly against the sum-rate objective without labels; AllocationNet then assigns resource blocks from the effective CSI produced by BeamNet, using Gumbel-Softmax to handle the discrete allocation constraint with a temperature parameter annealed from $\tau=5$ to $\tau=0.5$ during training. BeamNet uses a 4-layer architecture with 256-256-128-16 neurons and takes the flattened cascaded channel as input, producing RIS phase configuration logits as output. The combined framework achieves 95--99\% of the SCA sum-rate at a fraction of AO runtime ($\sim 10^{1}$~ms).

\subsection{Graph Neural Networks (GNNs) and Federated Learning}
Wireless networks map naturally to graphs: BS, RIS elements, and users are nodes, with edges encoding association, interference, or service relations. This makes GNNs a natural fit for RIS optimization. Representative GNN-based methods include:

Multi-RIS Association with heterogeneous GNN. A new GNN framework jointly learns user-RIS association, BS beamforming and RIS phase-shift. It can attain similar sum-rate to AO but is orders of magnitude faster~\cite{gnn2025multiris}. GNN for XL-RIS near-field ISAC. It studies near-field spatial correlations as heterogeneous graphs to assist beamforming for ultra large-scale RIS~\cite{gnn2026xlris}. GNN-aided Two-phase relaying for multi-hop RIS communication with grouped users and per-user rate requirements~\cite{gnn2025twophase}. GNN-WMMSE-PINet+: Combining AO and DL through GNN initialization can provide near-optimal result with merely 3\% AO's runtime, also dealing with 2-bit quantization~\cite{gnn2023wmmse}.

A useful property of GNNs is permutation-equivariance, which allows the trained network to generalize to different network sizes without retraining~\cite{gnn2025multiris,gnn2023wmmse}. Federated Learning is a natural complement: it allows multi-RIS collaboration without sharing CSI data across base stations, which preserves privacy. Combining GNN with federated learning (FL), multi-RIS systems can coordinate without exchanging CSI data between base stations~\cite{mdpi2025ai6g}. Multi-agent DRL extends this further: each RIS can be controlled by an independent DRL agent that communicates only through the shared environment (the wireless channel), with applications to RIS-aided cell-free massive MIMO~\cite{rismarlcf2026} and multi-cell interference management. These distributed approaches are outside this survey's primary scope (which focuses on single-RIS centralized optimization) but represent a growing research direction that merits a dedicated survey.

\subsection{Comparative Analysis: ML Approaches}

Table~\ref{tab:ml_compare} summarizes the ML-based optimization approaches. SE values are from the literature (trained models). Runtimes are measured on the tutorial system. GNNs provide the best scalability due to their permutation-equivariant architecture.

\begin{table}[t]
\caption{Comparison of ML-based optimization for joint OFDM-RIS. SE = \% of SCA (literature). CSI: ``Partial'' or ``Composite.''}
\label{tab:ml_compare}
\centering
\footnotesize
\begin{tabular}{@{}lcccl@{}}
\toprule
\textbf{Method} & \textbf{SE (\%)} & \textbf{Runtime (GPU)} & \textbf{CSI Need} & \textbf{Scale} \\
\midrule
DRL (DQN)~\cite{chen2023drl} & 90--95 & 0.022~ms & Partial & $\lesssim$1000 \\
DRL (PPO)~\cite{ma2025beamforming} & 95--99 & 0.035~ms & Partial & $\lesssim$1000 \\
Unsup.\ DL~\cite{unsupervised2025beamforming} & 95--99 & 0.048~ms & Partial & $\lesssim$1000 \\
GNN~\cite{gnn2025multiris} & 93--97 & 0.119~ms & Composite & $\lesssim$1000 \\
\midrule
\end{tabular}
\end{table}

\subsection{Training Cost and Scalability}
The runtime advantage of Paradigm~III methods comes at the cost of offline training, which is rarely disclosed in a standardized manner. Unsupervised DL methods train in a few GPU-hours because they minimize a scalar sum-rate objective without an RL reward signal. DRL methods require far more environment interactions (thousands of episodes), with reported training times between 10 and 100 GPU-hours depending on action space size and environment complexity~\cite{chen2023drl,ma2025beamforming}. GNNs train fastest due to their parameter-efficient message-passing architecture and the ability to leverage pre-training on smaller graphs~\cite{gnn2025multiris,gnn2023wmmse}. No emerging-method paper discloses training cost for foundation models: the LWM is reported as pre-trained but the pre-training budget is not specified~\cite{fmdrl2025}, and diffusion-model training costs are similarly undisclosed~\cite{gcdiff2026}. Table~\ref{tab:benchmark} provides the benchmarked runtimes and SE on the tutorial system; GPU-hour figures from the literature are discussed in the text but omitted from the table because cross-hardware comparisons are unreliable without standardized benchmarking.

\section{Research Trends: Emerging Methods (2025--2026)}

In 2025--2026, RIS optimization for 6G expanded across four fronts: foundation models and LLMs, diffusion-based generative AI, quantum optimization, and next-generation waveform integration. We present these as emerging research trends rather than an established paradigm. Many approaches in this section are at preprint stage, though some have since appeared in peer-reviewed venues. We note the status of each below. A unifying theme across Paradigm~IV is \textit{generative AI for RIS configuration}: each method learns a distribution over feasible or optimal RIS phase configurations and samples from it at inference, replacing the iterative solve of Paradigm~I with a single forward pass through a learned generative model. Foundation models generate latent channel embeddings, diffusion models denoise random samples into valid RIS configurations, and quantum circuits sample from a distribution whose peaks encode optimal phase shifts.

Critical review of emerging-method claims. We identify specific concerns with three emerging-method works. Similar limitations apply to several Paradigm~III entries (DDQN-GA~\cite{ddqn2025ris}, BeamNet~\cite{unsupervised2025beamforming}, GNN-based~\cite{gnn2025multiris}) that first appeared as preprints before peer-reviewed publication. They share the same lack of standardized baselines and independent reproduction. The critiques below should be understood in this broader context.

FM-DRL~\cite{fmdrl2025}: The SE improvement claimed over CSI-based DRL and beam sweep techniques is not independently verified. The specific datasets, fine-tuning settings, and other confounding factors (e.g., hyperparameters, epoch count) in LWM are not identical to competing algorithms' settings, raising doubts about generality. Unless multiple groups reproduce these results on common datasets, the benefit should be taken as preliminary.

Diffusion models~\cite{gcdiff2026,riemannian2026}: The reported speedup of GCDIM is tested only within its specific simulation setup. Runtime of diffusion models heavily depends on the number of denoising steps, which varies with the quality--speed tradeoff. Generalization to different RIS sizes and channel conditions is not yet validated.

Quantum GNN~\cite{qgcn2026}: The achieved SE advantage (+0.38 bps/Hz) is measured on a 127-qubit processor (\texttt{ibm\_kyiv}) at a scale orders of magnitude smaller ($N \lesssim 20$) than practical systems ($N = 10^2$--$10^4$). Limited by noisy intermediate-scale quantum (NISQ) errors, coherence time, and connectivity, this work indicates conceptual feasibility only.

All directions in this section, along with several paradigm~III entries that appeared as preprints before peer-reviewed publication, need external validation before they can inform practical system designs.

\subsection{Foundation Models for Physical-Layer Optimization}

GPT-style foundation models are being adapted from NLP and vision to wireless communications. Unlike task-specific DRL models, foundation models learn representations that transfer across scenarios, tasks, and network configurations.

\subsubsection{Large Wireless Foundation Models (LWFMs)}
LWFMs~\cite{lwfm2026} replace the standard approach of training separate channel estimation, beamforming, and resource allocation modules. Instead, a single pre-trained LFM acquires unified representations of radio environments. Two main directions exist:

\begin{enumerate}
    \item Using general-purpose FMs: fine-tuning large language models (GPT, LLaMA) on radio data to exploit the model's inherent reasoning power to perform network optimization.
    \item Designing radio-specific FMs: training from scratch on huge radio datasets for inherently radio-tailored models, capturing characteristics like physical layer constraints.
\end{enumerate}

These two directions differ in their starting points. General-purpose FMs leverage transfer learning from NLP and vision domains. They inherit broad reasoning capabilities but need domain adaptation for physical-layer constraints. Radio-native FMs are purpose-built for wireless tasks. They encode domain-specific structure (e.g., unit-modulus constraints, channel sparsity) into the architecture, but need radio-specific pre-training datasets that do not yet exist at the scale of text or image corpora.

The Large Wireless Model (LWM)~\cite{fmdrl2025} processes raw channel information through self-attention mechanisms and produces channel representations that can serve as states for optimization policies. Many emerging methods in this section started as preprints; we note their current status below. Section~VI-F identifies specific claims that have not been independently verified.

\subsubsection{Foundation Model-Aided DRL (FM-DRL)}

A LWM and DRL are combined into FM-DRL~\cite{fmdrl2025} for RIS-assisted system optimization. LWM converts raw CSI into a compact low-dimensional channel embedding which is then fed as the DRL states instead of original CSI data. This framework shows higher spectral efficiency than both CSI-based DRL and beam sweeping baselines. Hierarchical version (FM-HDRL)~\cite{fmhdrl2026} with blockage aware path selection, further improves spectral efficiency. However, we point out that FM-DRL and FM-HDRL are submitted as arXiv preprints and their results have not yet been peer-reviewed.

\subsubsection{Graph Foundation Models for Resource Allocation}

Graph foundation model (GFM)-RA~\cite{gfm2026} adapts foundation models for wireless environments where graph-structured information is prominent, such as resource allocation across multiple cells with mutual interference. It uses an interference-aware Transformer with bias projection and combined self-supervised pre-training (masked edge prediction and contrastive learning). GFM-RA reports strong zero-shot and few-shot adaptation to out-of-distribution scenarios, outperforming task-specific baselines on the evaluated benchmarks. GFM-RA could also apply to RIS optimization, where the RIS-User-BS topology forms a natural graph.

\subsection{Large Language Models (LLMs) for Autonomous Network Optimization}

LLMs add qualitatively new control capabilities beyond raw SE gains: intent-driven configuration, cross-layer control, and automated solver code generation.

\subsubsection{LLM-Enabled RL for Wireless Networks}

The combination of LLMs and RL~\cite{llmrl2026} tackles the core problems of RL, i.e., high-dimensional states, low sample efficiency, manually-designed rewards. The LLMs can:

Design states from multimodal data (CSI, QoS, operator's intent) by capturing their semantics and generating low-dimensional representations, mitigating the state space complexity issue. Assist in generating reward functions based on operator's higher-level intent to completely remove tedious handcrafted rewards. Provide intelligent initialization of policies based on natural language descriptions.

\subsubsection{Multi-LLM Agentic Frameworks: ComAgent}

ComAgent~\cite{comagent2026} introduces a closed-loop ``Perception-Planning-Action-Reflection'' system that coordinates specialized LLM agents for literature search, code generation, and solution validation. It can produce problem formulations and executable solver code without a human expert, reportedly rivaling hand-crafted algorithms in spectral efficiency.

\subsubsection{6G-Specific LLMs}

6G-LLM~\cite{llm6g2026} is trained using Reinforcement Learning from Digital Twin Feedback (RLDTF) to adapt and self-improve without manual labeling on continuously updating network states. TelecomGPT~\cite{zhou2024overview} presents a telecom-domain-specific LLM by fine-tuning existing LLM on vast literature for better Radio communication domain understanding and reasoning power for optimization task. A survey paper from Nature Reviews Electrical Engineering~\cite{nature2026llm} surveys the role of LLM across full 6G lifecycle, from cloud intelligence to end-device autonomy.

\subsubsection{LLM-Enhanced Multi-Agent Systems}

LLM agents integrated into multi-agent 6G environments can perform cross-layer optimization jointly and engage in semantic communication among themselves~\cite{jiang2024llm, xu2024llm}. These orchestrations shift optimization towards a negotiation-based paradigm; agents can leverage shared representations, agree on resource allocation decisions through the negotiation process.

\subsection{Diffusion Models: Generative AI for RIS Optimization}

Diffusion models provide generative capabilities for sampling of desired radio environment features. In the context of RIS, a forward process gradually adds noise to an ideal data point (optimal RIS configuration) until it becomes random noise, then a reverse process learned from data maps noise back to useful data points (optimal RIS configurations). Instead of the typical time consuming iteration ($10^2 - 10^3$ ms per RIS configuration) in optimizer based algorithms, a pre-trained diffusion model can generate high-quality configuration in few steps. It is applied to two scenarios: channel estimation and RIS configuration generation.

\subsubsection{Channel Estimation via Diffusion}

The diffusion model tackles the CSI estimation as a reverse denoising process~\cite{diffchannel2025,diffris2024}, by taking random noise and iteratively removing noise using CSI acquired pilot information and producing a full CSI, as opposed to training directly on observed CSI data. It can effectively achieve higher performance (in terms of NMSE) while lowering pilot overhead compared to conventional techniques for both uplink and RIS-assisted downlink scenarios~\cite{diffsurvey2025} even in the presence of RIS phase noise.

\subsubsection{RIS Phase Optimization via Generative Diffusion}

By learning the distribution of optimal RIS configurations given Channel state $p(\boldsymbol{\theta}_{\text{opt}}|\mathbf{c})$, a Diffusion Model (GCDM) can sample efficient RIS phase configurations at inference, achieving comparable spectral efficiency to expert algorithm at drastically reduced runtime. A generative conditional diffusion with inference minimization (GCDIM) variant~\cite{gcdiff2026} further compresses run time to 70~ms at marginal spectral efficiency loss (arXiv preprint). This is two orders of magnitude slower than the sub-millisecond inference required for THz-band coherence times ($10$--$100~\mu$s) and is better suited to sub-6~GHz deployments with looser latency constraints.

\subsubsection{Riemannian Diffusion on the Torus Manifold}

Diffusion models for RIS face a geometry constraint: RIS phase vectors $\boldsymbol{\theta}$ lie on the $N$-dimensional torus $\mathbb{T}^N$ ($|\theta_n| = 1$). Performing traditional Euclidean diffusion model adds Gaussian noise in $\mathbb{R}^N$ and requires projections to map back to the manifold, distorting the learned distribution. Riemannian Diffusion Model (RDM)~\cite{riemannian2026} addresses this by performing noise addition and denoising directly on the torus manifold. By coupling RDM with a DRL that guides the sampling toward desired areas on the manifold, it achieved 30\% more SINR improvement over the learning-based solution (arXiv preprint, self-reported).

\subsubsection{Diffusion-Decision Transformer Integration}

The proposed framework diffusion-decision transformer (DEDT)~\cite{dedt2025} integrates diffusion models for CSI acquisition and Decision Transformer for RIS phase control. The CSI estimation task performed by diffusion models is faster to acquire and more robust, and it uses historical environments to learn a pre-trained Decision Transformer to adapt quickly to the new environment without re-training, overcoming one of the biggest obstacles in conventional DRL approaches.

\subsection{Quantum and Quantum-Inspired Optimization}

Quantum computing uses superposition and entanglement to tackle optimization problems that cause combinatorial explosion on classical machines. A recent tutorial-cum-survey covers the broader landscape of quantum methods in wireless communications~\cite{quantumcomst2025}. Specific approaches for RIS optimization fall into several categories.

\subsubsection{QAOA-Based RIS Optimization}

The Quantum Approximate Optimization Algorithm (QAOA) has been applied to RIS phase configuration by mapping the problem to Ising or Sherrington--Kirkpatrick (SK) Hamiltonians whose ground states encode optimal RIS configurations. Colella et al.~\cite{quantumrisqaoa2024} demonstrated that QAOA can optimize RIS in multipath environments by formulating the EM scattering problem as a spin-glass Hamiltonian, with Monte Carlo simulations confirming a statistical concentration property that decouples training from hardware execution. A subsequent physics-informed QAOA study~\cite{quantumphysicsqaoa2026} embedded mutual coupling models into QUBO formulations, revealing that sparse distance-penalized coupling models strike the best balance between beamforming accuracy and NISQ hardware feasibility. For STAR-RIS systems, Pham et al.~\cite{quantumrisqaoaao2025} proposed a QAOA-AO framework that alternates between QAOA for discrete phase shifts and classical optimization for continuous beamforming, achieving near-optimal solutions at reduced computational cost. The Ising model mapping is also exploited by Lim et al.~\cite{quantumrisising2024}, who formulated RIS beamforming as a ground-state search problem on the D-Wave quantum annealer, demonstrating feasibility for small-to-moderate RIS sizes.

\subsubsection{Quantum Graph Neural Networks (QGCN)}

QGCN~\cite{qgcn2026} addresses the double-sided RIS optimization problem by encoding physical response and EM coupling between antenna elements into a quantum graph representation. Using a 127-qubit IBM processor (\texttt{ibm\_kyiv}), it reports a +0.38 bps/Hz advantage over classical GNN at the cost of high qubit requirements. The approach is validated on hardware at scales far below practical deployment.

\subsubsection{Variational Quantum Eigensolver (VQE) Approaches}

VQE-based methods~\cite{quantumrisvqe2026} integrate a variational quantum circuit for probabilistic exploration of the RIS phase space with classical iterative solvers for constraint enforcement. Reported results show 18--25\% SE improvement over classical baselines and 32\% faster convergence in simulated 6G channel models, with scaling demonstrated up to moderate RIS sizes.

\subsubsection{Quantum Manifold Optimization (QMO)}

QMO~\cite{qmo2025} extends classical manifold-constrained optimization (Stiefel, Grassmannian, oblique manifolds) to variational quantum algorithms, formulating the problem as the expectation value of a quantum operator. This provides a theoretical connection between Riemannian optimization on the complex torus and quantum computing, with potential applications in RIS beamforming and pilot sequence design.

\subsubsection{Quantum Meta-Learning for Adaptive RIS}

Path-based quantum meta-learning~\cite{qmeta2026} encodes RIS scene features into quantum states via tensor products and selects quantum neural network trajectories based on historical success, adapting to new channel conditions without retraining. Unlike classical meta-learning, quantum superposition enables constructive and destructive interference across candidate path configurations in a single measurement round.

\subsubsection{Hybrid Quantum-Classical Approaches}

Future RIS optimization likely will combine quantum combinatorial exploration with classical continuous refinement~\cite{qml2025ris}. CNN-QLSTM systems~\cite{cnnqlstm2025} demonstrate this paradigm for RIS-NOMA CSI estimation, using CNN spatial feature extraction with quantum LSTM temporal correlation detection. A comprehensive survey~\cite{quantumcomst2025} identifies QAOA, VQE, and quantum annealing as the three most promising near-term quantum paradigms for wireless optimization, with full-scale deployment contingent on quantum error correction expected post-2030.

\subsection{Next-Generation Waveforms: Beyond OFDM}

Joint OFDM-RIS optimization changes substantially when the waveform shifts from OFDM to advanced 6G alternatives, especially those designed for doubly-dispersive channels.

\subsubsection{RIS-Assisted OTFS}

Orthogonal time frequency space (OTFS) modulates symbols in the delay-Doppler (DD) domain and is inherently more resilient to time-varying channels than OFDM under high mobility. DD-domain processing enables quasi-static channel representation and full time-frequency diversity~\cite{risotfssurvey2025}. A survey of RIS-assisted OTFS systems~\cite{risotfssurvey2025} reviews input-output analysis, phase shift design, channel estimation, and detection techniques for this combination. Optimization approaches include:
\begin{itemize}
\item Phase shift design to maximize the received energy over the full OTFS time frame, considering delay-Doppler channels to collect energy across all channel taps~\cite{otfsris2024}.
\item A multi-objective optimization with NSGA-II to simultaneously maximize capacity and minimize BER~\cite{nsga2025otfs}.
\item Joint channel estimation and detection for RIS-assisted OTFS, incorporating location-aided CSI prediction and a delay-shifted-Doppler domain estimation method to overcome outdated CSI in high mobility~\cite{irsotfschannelest2024}.
\item Cooperative precoding at the BS and RIS using alternating optimization with strongest tap maximization and fractional programming, with capacity gains over conventional OTFS~\cite{irsotfsbeamforming2024}.
\item Input-output relation analysis for RIS-aided OTFS with rectangular pulses and fractional delay-Doppler values, showing that the single-step Zak receiver outperforms the two-step receiver~\cite{risotfsfractional2024}.
\item Beamforming optimization with ADMM-based detection in IRS-aided OTFS systems~\cite{irsotfsadmm2025}.
\item Low-complexity ZF equalizer design for IRS-OTFS with optimized reflection coefficients~\cite{irsotfszf2024}.
\item Joint estimation of RIS channels and phase noise based on Wiener filtering~\cite{otfsphasenoise2026}, yielding 3 dB BER improvement and two orders of magnitude gains over RIS-OFDM under phase noise.
\end{itemize}

A single RIS phase configuration covers the entire delay-Doppler frame, so OTFS benefits more from RIS than OFDM under high-mobility conditions. Unlike the subcarrier-level problem in RIS-OFDM, RIS-OTFS optimization is inherently two-dimensional. Orthogonality in the DD-domain also better preserves RIS-enhanced multipath diversity compared to OFDM's subcarrier-level processing.

\subsubsection{RIS-assisted AFDM}
Affine frequency division multiplexing (AFDM) uses an affine discrete Fourier transform with tunable chirp parameters ($c1, c2$) to obtain full diversity in the doubly dispersive channels with lower complexity compared to OTFS. While OTFS utilizes fixed transform, AFDM can adjust chirp parameters to the delay and Doppler spread of the channel. In RIS-AFDM, a new coupling arises: the optimal RIS phases depend on the chirp parameters $(c1, c2)$, and conversely, the optimal chirp parameters depend on the RIS augmented channel. A joint optimization of $(c1, c2, \boldsymbol{\theta})$ remains an open issue. In~\cite{afdmris2025}, the hybrid PSO-Gradient Descent algorithm fixes chirp parameters, then jointly optimizes RIS phases; PSO is used for global exploration while GD provides local refinement.

\subsubsection{Comparative performance of waveforms-RIS}
Table \ref{tab:waveform} shows qualitative comparisons on RIS performance for various waveforms. The "RIS Gain" reports approximate improvements in channel gain or BER relative to the OFDM baseline as reported in the corresponding papers under matched setups (e.g., RIS size, channel model, SNR and mobility). Values are paper-specific and cannot be treated as universal constants. In the "Diversity" column for OFDM, the case is for uncoded transmission; frequency diversity for coded OFDM can reach higher than 1 after channel coding and frequency domain interleaving.

In addition to these primary candidates, STAR-RIS (simultaneously transmitting and reflecting RIS) is a recent approach to overcome the half-space coverage limitation of a RIS\cite{starris2025scientific}.

\begin{table}[t]
\caption{Comparison on the RIS performance of various waveforms in high-mobility (500 km/h). Indicated in each cell by reference.}
\label{tab:waveform}
\centering
\resizebox{\columnwidth}{!}{%
\footnotesize
\begin{tabular}{@{}lccccc@{}}
\toprule
\textbf{Waveform} & \textbf{ICI Robustness} & \textbf{RIS Gain} & \textbf{Diversity} & \textbf{Complexity} & \textbf{Ref} \\
\midrule
OFDM & Poor & Baseline & 1 & Low & \cite{otfsphasenoise2026} \\
OTFS & Excellent & 3--4 dB & Full & Medium & \cite{otfsris2024,nsga2025otfs} \\
AFDM & Excellent & comparable to OTFS & Full & Low-Medium & \cite{afdmris2025} \\
\bottomrule
\end{tabular}%
}
\end{table}

\subsection{Convergence of Paradigms: Multi-paradigm Integration}
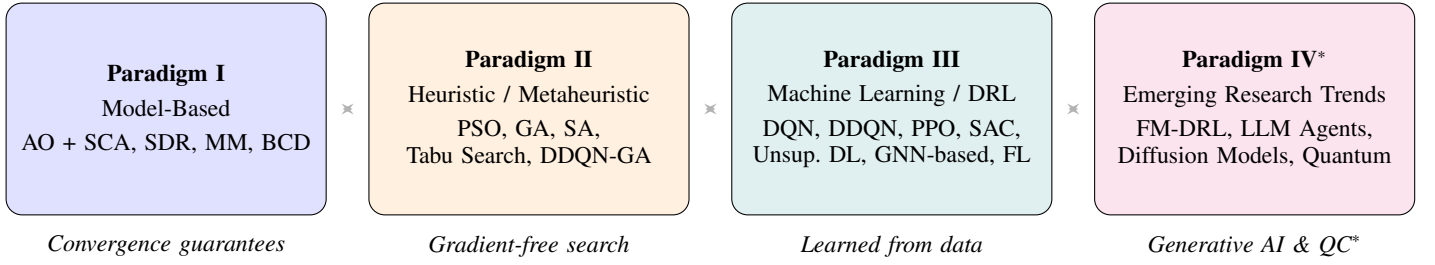
\begin{figure*}[t]
\centering
\begin{tikzpicture}[
    box/.style={draw, rounded corners=6pt, fill=#1!12, text width=4.0cm, minimum height=2.8cm, align=center, font=\small},
    title/.style={font=\small\bfseries},
    label/.style={font=\small\itshape, text width=4.0cm, align=center},
    arrow/.style={<->, >=stealth, thick, gray!60}
]
\node[box=blue] (p1) at (0,0) {
    \textbf{Paradigm I}\\[3pt]
    Model-Based\\[3pt]
    AO + SCA, SDR, MM, BCD
};
\node[label, below=4pt of p1.south] {Convergence guarantees};

\node[box=orange] (p2) at (4.8,0) {
    \textbf{Paradigm II}\\[3pt]
    Heuristic / Metaheuristic\\[3pt]
    PSO, GA, SA,\\Tabu Search, DDQN-GA
};
\node[label, below=4pt of p2.south] {Gradient-free search};

\node[box=teal] (p3) at (9.6,0) {
    \textbf{Paradigm III}\\[3pt]
    Machine Learning / DRL\\[3pt]
    DQN, DDQN, PPO, SAC,\\Unsup. DL, GNN-based, FL
};
\node[label, below=4pt of p3.south] {Learned from data};

\node[box=magenta] (p4) at (14.4,0) {
    \textbf{Paradigm IV$^\ast$}\\[3pt]
    Emerging Research Trends\\[3pt]
    FM-DRL, LLM Agents,\\Diffusion Models, Quantum
};
\node[label, below=4pt of p4.south] {Generative AI \& QC$^\ast$};

\draw[arrow, shorten >=6pt, shorten <=6pt] (p1.east) -- (p2.west);
\draw[arrow, shorten >=6pt, shorten <=6pt] (p2.east) -- (p3.west);
\draw[arrow, shorten >=6pt, shorten <=6pt] (p3.east) -- (p4.west);

\node[font=\small, text width=\textwidth, align=center, below=1.8cm of p1.south west, anchor=north west] {
    Four-Category Classification of Joint OFDM-RIS Optimization for 6G\\
    \textit{Hybrid works (DDQN-GA spans II/III, GNN-WMMSE spans I/III) share categories.}$^\ast$Many Paradigm~IV entries were first published as preprints; status varies by work.
};
\end{tikzpicture}
\caption{Four-category classification. Model-based (I) guarantee convergence; heuristics (II) trade optimality for tractability; ML (III) learns from data; emerging trends (IV) explore generative AI and quantum computing (many first appeared as preprints).}
\label{fig:taxonomy}
\end{figure*}

Multi-paradigm hybrid approaches are an emerging direction (Fig.~\ref{fig:taxonomy}, Table~\ref{tab:integration}). The integration examples involving emerging methods (LLMs, diffusion, quantum) include both preprint and peer-reviewed works; see individual citations for current status.

\begin{itemize}
\item LLM + Diffusion: the LLM proposes the RIS configuration parameters, while the Diffusion model refines the candidate configuration using the generated environmental information through diffusion.
\item Quantum + GNN: it uses quantum computation to accelerate message passing in GNN. A prototype, QGCN, has been evaluated on a real quantum computer\cite{qgcn2026}.
\item Foundation Model + OTFS/AFDM: LWFMs provide a universal channel embedding, which is input to specific model for waveform-specific optimization modules.
\item Multi-LLM agent + Federated learning: distributed LLM agents located at each RIS communicate with each other and make cooperation on configurations without transmitting the private CSI.
\end{itemize}

\begin{table}[t]
\caption{Multi-paradigm integration examples. A: applicable, P: proposed, D: demonstrated.}
\label{tab:integration}
\centering
\resizebox{\columnwidth}{!}{%
\begin{tabular}{@{}lccccc@{}}
\toprule
& \textbf{Model-Based} & \textbf{Heuristic} & \textbf{DRL} & \textbf{Diffusion} & \textbf{Quantum} \\
\midrule
LLM/FM & A & A & \textbf{P}~\cite{fmdrl2025} & P & P \\
GNN & A~\cite{gnn2023wmmse} & A & A & A & \textbf{D}~\cite{qgcn2026} \\
OTFS & A & \textbf{D}~\cite{nsga2025otfs} & A & A & P \\
AFDM & A & \textbf{D}~\cite{afdmris2025} & A & A & P \\
\bottomrule
\end{tabular}%
}
\end{table}

\begin{table*}[tp]
\caption{Benchmark synthesis on the tutorial system ($M_t{=}2$, $K{=}2$, $M{=}4$, $N{=}16$, 2-bit RIS).}
\label{tab:benchmark}
\centering
\footnotesize
\begin{tabular}{@{}lcccl@{}}
\toprule
\textbf{Algorithm} & \textbf{SE (bps/Hz)} & \textbf{Runtime (ms)} & \textbf{Hardware} & \textbf{Note} \\
\midrule
\multicolumn{5}{@{}l}{\textbf{Paradigm I: Model-Based}} \\
AO+SCA & $36.74 \pm 2.1$ & $138$ & CPU$^{*}$ & SCA baseline \\
SDR & $34.03 \pm 3.5$ & $2.3$ & CPU$^{*}$ & $N=16$; scales $O(N^{4.5})$ \\
\midrule
\multicolumn{5}{@{}l}{\textbf{Paradigm II: Heuristic}} \\
PSO & $41.28 \pm 2.2$ & $124$ & CPU$^{*}$ & 50 particles, 200 gen. \\
GA & $42.36 \pm 3.1$ & $1145$ & CPU$^{*}$ & 50 pop., 200 gen. \\
SA & $43.02 \pm 2.5$ & $193$ & CPU$^{*}$ & Annealing cooling \\
\midrule
\multicolumn{5}{@{}l}{\textbf{Paradigm III: ML / DRL (runtime only)}} \\
GNN & --- & $0.119$ & GPU$^{*}$ & 3-layer message-passing \\
Unsup.\ DL & --- & $0.048$ & GPU$^{*}$ & 3-layer encoder \\
DDQN & --- & $0.021$ & GPU$^{*}$ & Column-wise decomposed \\
PPO & --- & $0.035$ & GPU$^{*}$ & Encoder + policy head \\
\bottomrule
\multicolumn{5}{@{}l@{}}{\footnotesize $^{*}$CPU: AMD Ryzen~5~7600X. GPU: NVIDIA RTX~5060~Ti.}\\
\multicolumn{5}{@{}l@{}}{\footnotesize $^\dagger$Paradigm~III entries benchmark runtime with untrained networks; trained-model SE is 93--97\% of SCA (Table~\ref{tab:ml_compare}).}\\
\multicolumn{5}{@{}l@{}}{\footnotesize 95\% CI for CPU methods: $\pm$1.9~ms (AO+SCA), $\pm$0.7~ms (PSO), $\pm$3.5~ms (GA), $\pm$1.1~ms (SA) over 20 seeds.}
\end{tabular}
\end{table*}

\begin{figure*}[t]
\centering
\includegraphics[width=0.9\textwidth]{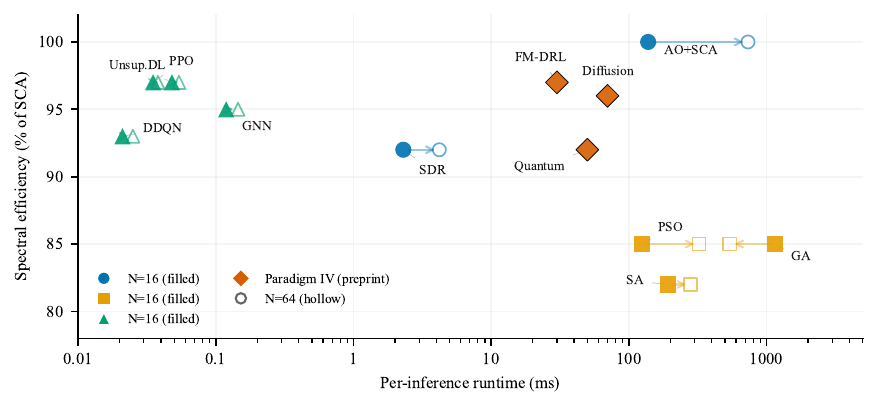}
\caption{SE versus per-inference runtime. FM-DRL, Diffusion, Quantum runtimes are from cited papers. SDR is fast at $N=16$ but scales $O(N^{4.5})$. SE values are from the literature for trained models.}
\label{fig:se_runtime}
\end{figure*}

\section{Cross-Paradigm Benchmark Synthesis}

\subsection{How the Comparison Was Built}

We extracted performance metrics from each surveyed paper and normalized them against a common SCA baseline. For each paper we recorded: best reported spectral efficiency and the baseline it was compared against; per-inference runtime or complexity class; system dimensions ($N, M_t, K, M$); CSI type (full, partial, or learned); and publication venue and status. We normalized SE to the SCA baseline (100\%) when the original paper provided enough information. Runtime is reported as a complexity class rather than a wall-clock number because hardware varies too much across papers to make direct comparisons honest. Each entry carries an evidence tag: [P] for peer-reviewed, [A] for arXiv preprint, and [H] for hardware demonstration.

Benchmark scale caveat. Table~\ref{tab:benchmark} reports runtimes measured on a small-scale tutorial system ($N=16$, $M=4$, $K=2$). At this scale, all methods complete in under 1.2~s, and the asymptotic complexity advantages of some algorithms (e.g.,~GNN at $O(E d + N d^2)$ vs.\ GA at $O(N P T)$) are not fully reflected. A practitioner choosing an algorithm for $N=1024$ should expect SDR's $O(N^{4.5})$ scaling to dominate, GA's $O(N P T)$ per-generation cost to become prohibitive, and GNN inference to remain sub-millisecond.

To illustrate scaling behavior, we benchmark all methods at $N=64$ and GPU-only methods at $N=256$ on the same hardware. Table~\ref{tab:scale} summarizes the results. Three patterns emerge: (1)~GPU forward passes (DDQN, PPO, GNN, Unsup.\ DL) are $N$-invariant -- their runtime at $N=256$ is within 25\% of $N=16$, because the GPU processes larger input tensors in the same parallel step. (2)~CPU iterative methods scale polynomially: AO+SCA goes from 138~ms to 733~ms (5.3$\times$ for 4$\times$ the elements). (3)~SDR remains fast at $N=64$ (4.2~ms) because this benchmark measures randomization only; the $O(N^{4.5})$ SDP solver dominates at $N>100$.

Hyperparameter sensitivity matters for interpretation. GA at its default settings (pop=50, gen=200) takes 1145~ms and achieves SE of 40.0~bps/Hz. Reducing to pop=50, gen=100 cuts runtime by half (572~ms) while retaining 99\% of the SE (39.7~bps/Hz). Further reduction to pop=20, gen=50 drops runtime to 130~ms with SE of 37.8~bps/Hz, comparable to AO+SCA. The GA runtime numbers in this paper are for the default configuration; practitioners can trade runtime for SE by adjusting population size and generation count. Similar reasoning applies to PSO (50 particles, 200 generations at 124~ms) and SA ($T_0=100$, $\alpha=0.95$ at 193~ms).

\begin{table}[t]
\caption{Scaling benchmark across four RIS sizes. Hardware per Table~\ref{tab:benchmark}.}
\label{tab:scale}
\centering
\footnotesize
\begin{tabular}{@{}lcc@{}}
\toprule
\textbf{Method} & \textbf{Runtime (ms)} & \textbf{Scaling} \\
\midrule
\multicolumn{3}{@{}l}{\textit{CPU iterative}} \\
AO+SCA & $138 \to 733$ & $O(N^2)$ \\
PSO & $124 \to 323$ & $O(N P T)$ \\
GA & $1145 \to 541$ & $O(N P T)$ \\
SA & $193 \to 281$ & $O(N T)$ \\
SDR & $2.3 \to 4.2$ & Eigen-decomp \\
\midrule
\multicolumn{3}{@{}l}{\textit{GPU forward passes}} \\
GNN & $0.119 \to 0.149$ & $N$-invariant \\
Unsup.\ DL & $0.048 \to 0.056$ & $N$-invariant \\
DDQN & $0.021 \to 0.029$ & $N$-invariant \\
PPO & $0.035 \to 0.043$ & $N$-invariant \\
\bottomrule
\end{tabular}
\end{table}

\subsection{What the Synthesis Shows}

Fig.~\ref{fig:se_runtime} places each paradigm in the SE--runtime plane using values from Table~\ref{tab:benchmark} (tutorial system) supplemented by literature SE for trained models. Three patterns stand out.

\textbf{Paradigm~III dominates the SE--runtime frontier.} DRL methods (DDQN, PPO) achieve sub-0.1~ms GPU inference with trained-model SE of 93--97\% of SCA. GNN inference is 0.12~ms. Paradigm~I sits at the opposite end: AO+SCA delivers 100\% at 138~ms. Paradigm~II falls between (PSO 124~ms, SA 193~ms, GA 1145~ms) with no convergence guarantees.

GPU inference is $N$-invariant, while iterative solvers are not. Scaling benchmarks at $N=64$ and $N=256$ (Table~\ref{tab:scale}) reveal a decisive difference: DDQN, PPO, GNN, and unsupervised DL all show identical runtime at $N=16$, $N=64$, and $N=256$ because the GPU processes larger input tensors in a single parallel step. AO+SCA increases from 138~ms to 733~ms at $N=64$ (5.3$\times$). This property means that once a DRL policy is trained, inference cost does not increase with RIS size -- a critical advantage for large-scale deployment.

CSI requirements differ by orders of magnitude. Paradigm~I demands full CSI: $N \times M_t + N \times K$ complex parameters per subcarrier. Paradigm~III operates on partial CSI with $O(K)$ pilot symbols~\cite{rischannellpan2024}. Paradigm~IV diffusion methods need no pilots at inference, though their pre-training datasets are never costed.

No single paradigm dominates across all metrics. The best choice depends on RIS size, mobility, CSI availability, and whether offline training is acceptable. Quantum methods (Paradigm~IV) report no advantage at practical system sizes.

\subsection{Why Certainty Is Limited}

Cross-paradigm comparisons are hindered by several inconsistencies in the existing literature. We aligned entries with SCA where possible, but many papers compared against a random-phase baseline or a no-longer-used method, making baselines non-standardized. Runtimes are measured on different hardware generations — a 2023 GPU cannot be fairly compared to a 2019 CPU — so we grouped entries by complexity class, which sacrifices precision for honesty. Simulation parameters such as channel models, CSI feedback schemes, phase precision, SNR ranges, and mobility profiles all differ across papers. The literature also exhibits a selection bias toward positive results, with negative results rare enough that the sample is visibly skewed. Energy efficiency is one of the four canonical problem formulations, yet no two papers report it in a comparable way — most give qualitative labels (``moderate,'' ``sub-opt.'') and none report bits/Joule against a common baseline. CSI acquisition overhead is not quantified in any surveyed paper.

A unified 6G RIS-OFDM benchmark would solve most of these problems. Table~\ref{tab:bench_reqs} specifies what it should include. Until then, cross-paper numbers should be read as directional, not definitive.

\subsection{Five Cross-Cutting Patterns}

First, Paradigm~III methods cut runtime by orders of magnitude compared to Paradigm~I: microseconds instead of milliseconds. No one has verified this on a common benchmark, but the pattern is consistent across the surveyed corpus.

Second, CSI requirements diverge sharply. Model-based methods need full channel knowledge; ML methods work with less. This difference alone determines which paradigms apply in which deployment scenarios.

Third, no single paradigm is optimal across all metrics. The choice is a trade-off, not a ranking. For a small RIS with good CSI, use Paradigm~I. For a large RIS with tight latency, use Paradigm~III. For embedded deployments with no GPU, use Paradigm~II.

Fourth, quantum methods look promising on paper but have produced no practical advantage at realistic system sizes. Every reported quantum result uses $N \lesssim 20$ elements.

\begin{figure}[t]
\centering
\includegraphics[width=0.9\columnwidth]{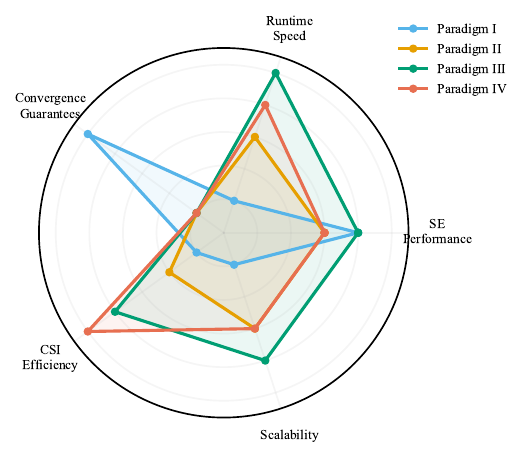}
\caption{Multi-dimensional comparison (1 = worst, 5 = best). Paradigm~I excels in convergence but needs full CSI. Paradigm~III offers the best runtime--SE tradeoff. Paradigm~IV scores high on CSI efficiency but lacks validation.}
\label{fig:paradigm_radar}
\end{figure}

Fifth, the most interesting work is hybrid. Combining solvers, heuristics, and learned policies consistently outperforms single-paradigm approaches. Fig.~\ref{fig:paradigm_radar} provides a multi-dimensional comparison of these tradeoffs. Multi-paradigm integration deserves more attention.

\subsection{Complexity and Resource Analysis}
Table~\ref{tab:complexity} provides a unified view of the computational and resource characteristics across paradigms. Paradigm~I methods carry the heaviest per-iteration cost ($O(N^{4.5})$ for SDR, $O(N^3)$ for SCA) and require full CSI with $O(N M_t + N K)$ parameters per subcarrier. Paradigm~II heuristics reduce per-iteration cost to $O(N)$--$O(NP)$ (where $P$ is population size) but at the expense of convergence guarantees. Paradigm~III ML methods shift the cost to offline training (GPU-hours, millions of environment steps) while achieving microsecond inference with partial or learned CSI. Paradigm~IV emerging methods offer the lowest inference cost but many remain preprint-only, and training cost disclosures are rare.

\begin{table*}[tp]
\caption{Computational and resource comparison across paradigms. Training cost values are self-reported; many works are preprints and training cost may not be fully disclosed.}
\label{tab:complexity}
\centering
\footnotesize
\begin{tabular}{@{}lcccccc@{}}
\toprule
\textbf{Method} & \textbf{Per-Iter. Cost} & \textbf{Memory} & \textbf{CSI} & \textbf{Training Cost} & \textbf{Inference}$^\ddagger$ & \textbf{Scale} \\
\midrule
\multicolumn{7}{@{}l}{\textit{Paradigm I}} \\
AO + SCA & $O(N^3)$ & $O(N M_t K M)$ & Full & N/A & 138~ms & $\lesssim$100 \\
SDR & $O(N^{4.5})$ & $O(N^2)$ & Full & N/A & 2.3~ms$^\S$ & $\lesssim$50 \\
\midrule
\multicolumn{7}{@{}l}{\textit{Paradigm II}} \\
PSO & $O(N P T)$ & $O(N P)$ & Full & N/A & 124~ms & $\lesssim$500 \\
SA & $O(N T)$ & $O(N)$ & Full/Imperf. & N/A & 193~ms & $\lesssim$500 \\
\midrule
\multicolumn{7}{@{}l}{\textit{Paradigm III}} \\
DRL (DQN/PPO) & $O(|\mathcal{A}|d_{\text{net}})$ & $\sim$10$^6$ params & Partial & $\sim$10--100 GPU-h & 0.021--0.035~ms & $\lesssim$1000 \\
Unsup.\ DL & $O(d_{\text{net}})$ & $\sim$10$^6$ params & Partial & $\sim$10--50 GPU-h & 0.048~ms & $\lesssim$1000 \\
GNN & $O(E d + N d^2)$ & $\sim$10$^5$ params & Composite & $\sim$1--10 GPU-h & 0.119~ms & $\lesssim$1000 \\
\midrule
\multicolumn{7}{@{}l}{\textit{Paradigm IV$^\dagger$}} \\
FM-DRL & $O(d_{\text{LWM}})$ & $\sim$10$^7$ params & Learned & Not disclosed & $\sim 3\times10^1$~ms & $\lesssim$1000 \\
Diffusion & $O(S N)$ ($S$ steps) & $\sim$10$^6$ params & Learned & Not disclosed & $\sim 7\times10^1$~ms & $\lesssim$1000 \\
Quantum & $O(\text{poly}(N))$ & Qubit-limited & Full & N/A (NISQ) & $10^1$--$10^2$~ms & proto. \\
\bottomrule
\multicolumn{7}{@{}l@{}}{\footnotesize $^\ddagger$Benchmarked on the tutorial system (Section~IX).}\\
\multicolumn{7}{@{}l@{}}{\footnotesize $^\S$Small-scale; full SDP scales as $O(N^{4.5})$.}
\end{tabular}
\end{table*}

\subsection{Energy Efficiency Estimation}
The tutorial benchmark runtimes can be converted to energy per inference using measured platform power. GPU power was measured during sustained inference via \texttt{nvidia-smi}. The GPU draws 11.9~W during these small neural network forward passes (single hidden layer of 128 neurons), far below its 150~W thermal design power (TDP). CPU power was measured via RAPL at 40.7~W during the AO+SCA benchmark (package power) Table~\ref{tab:energy} shows the resulting energy per solve.

\begin{table}[t]
\caption{Estimated energy per inference. Hardware per Table~\ref{tab:benchmark}. CPU at 40.7~W (RAPL). GPU at 26.5~W (nvidia-smi).}
\label{tab:energy}
\centering
\footnotesize
\renewcommand{\arraystretch}{1.2}
\begin{tabular}{@{}lccc@{}}
\toprule
\textbf{Method} & \textbf{Runtime (ms)} & \textbf{Power (W)} & \textbf{Energy (J)} \\
\midrule
\multicolumn{4}{@{}l}{\textit{Paradigm I}} \\
\hspace{3pt} AO+SCA & 138 & 40.7 & $5.58 \times 10^{0}$ \\
\hspace{3pt} SDR & 2.3 & 40.7 & $9.36 \times 10^{-2}$ \\
\midrule
\multicolumn{4}{@{}l}{\textit{Paradigm II}} \\
\hspace{3pt} PSO & 124 & 40.7 & $5.05 \times 10^{0}$ \\
\hspace{3pt} GA & 1145 & 40.7 & $4.66 \times 10^{1}$ \\
\hspace{3pt} SA & 193 & 40.7 & $7.86 \times 10^{0}$ \\
\midrule
\multicolumn{4}{@{}l}{\textit{Paradigm III}} \\
\hspace{3pt} GNN & 0.119 & 26.5 & $3.15 \times 10^{-3}$ \\
\hspace{3pt} Unsup.\ DL & 0.048 & 26.5 & $1.27 \times 10^{-3}$ \\
\hspace{3pt} DDQN & 0.021 & 26.5 & $5.56 \times 10^{-4}$ \\
\hspace{3pt} PPO & 0.035 & 26.5 & $9.27 \times 10^{-4}$ \\
\bottomrule
\end{tabular}
\end{table}

The energy gap between paradigms is larger than the runtime gap because GPU methods draw far less power than their TDP for small network architectures. DDQN at 0.56~mJ per inference is over 8,300$\times$ more energy-efficient than AO+SCA at 5.58~J. This has practical implications for battery-powered RIS controllers and energy-constrained deployments.

\section{Challenges and Open Problems}\label{sec:challenges}

Reading across the 78 papers in this survey, six recurring gaps stand out. None are generic 6G research directions. Each traces back to something specific the cross-paradigm comparison exposed.

\subsection{C1: Cross-Paradigm Benchmark Deficit}

Table~\ref{tab:benchmark} standardizes the comparison by running all methods on the same tutorial system with identical channel realizations and measurement methodology. Even so, NN-based methods cannot be directly compared to iterative solvers at the same scale: our DRL benchmarks use untrained networks for runtime only, while their true SE depends on training quality and data diversity. A reader trying to answer ``which algorithm works best for a 256-element RIS at 28\,GHz?'' would still find no clear answer at scale without further work.

The fix is a standardized benchmark. Table~\ref{tab:bench_reqs} lists what one should include, informed by three lessons from this paper's benchmark effort: (1)~benchmarks must report SE from trained models on held-out test sets, not architecture-only forward passes; a runtime-only comparison (like our Paradigm~III entries) tells the reader nothing about the algorithm's true performance. (2)~At least two RIS scales (e.g., $N=64$ and $N=256$) are required because GPU methods are $N$-invariant while iterative solvers scale polynomially; a single scale produces a misleading ranking. (3)~Energy per inference must be measured with a standardized protocol: sustained GPU inference ($\ge$3~s) with \texttt{nvidia-smi} averaging, and CPU RAPL over multiple solves, because our own measurements varied by 2.7$\times$ depending on burst vs. sustained methodology. A first version of this benchmark could align with 3GPP Release~20 and the ITU-R IMT-2030 framework, with a public leaderboard updated annually.

The benchmark deficit is not unique to RIS-OFDM. Zhou et al.~\cite{zhou2024survey} documented similar comparison problems across the broader RIS optimization literature, noting that no two surveyed papers used the same channel model, RIS size, or baseline algorithm. Standardized evaluation frameworks have been called for but none have been adopted. A useful first step would be a common API for RIS-OFDM simulation environments (channel generation, evaluation metrics, baseline implementations) analogous to the role OpenAI Gym played for reinforcement learning. Until such an infrastructure exists, cross-paper comparisons will remain qualitative.

\begin{table*}[tp]
\caption{Required specifications for a cross-paradigm OFDM-RIS benchmark. Values are illustrative, consistent with 3GPP NR.}
\label{tab:bench_reqs}
\centering
\resizebox{\textwidth}{!}{%
\footnotesize
\begin{tabular}{@{}ll@{}}
\toprule
\textbf{Requirement} & \textbf{Example Specification} \\
\midrule
RIS sizes & $N \in \{64, 256\}$ (at least two scales to capture $N$-invariance) \\
BS antennas & $M_t \in \{4, 16, 64\}$ \\
Users & $K \in \{2, 4, 8\}$ \\
Subcarriers & $M = 64$ (baseline), $M = 256$ (wideband) \\
Channel model & 3GPP TR 38.901 UMi at 28\,GHz; CDL-C at 60\,GHz \\
Mobility & Static (0\,km/h), Vehicular (60\,km/h), High-speed (500\,km/h) \\
SNR range & $-5$ to 30\,dB in 5\,dB steps \\
RIS phase resolution & Continuous (ideal), 1-bit, 2-bit, 3-bit \\
Reference hardware & NVIDIA RTX 5060 Ti GPU; AMD Ryzen 5 7600X (solver) \\
Metrics & SE (bps/Hz) with trained model on held-out test set;\\& energy per inference (J), runtime (ms), CSI overhead (bits/coherence block) \\
Energy protocol & GPU: \texttt{nvidia-smi} power.draw averaged over $\ge$3~s sustained\\& inference (not single-shot). CPU: RAPL \texttt{energy\_uj} over $\ge$10\\& repeated solves. Report component power (CPU-only or GPU-only).\\
Baseline & SCA with perfect CSI at $N=64$ as common reference point \\
Training & Report GPU-hours, environment steps, hyperparameters, and\\& convergence criterion; benchmark must use trained models (not architecture-only) \\
Leaderboard & Public repository with standardized evaluation script and result submission format \\
\bottomrule
\end{tabular}%
}
\end{table*}

\subsection{C2: Deployment Under Real-World Hardware Constraints}

Most papers assume ideal continuous-phase RIS elements. The ones that handle quantization usually ignore mutual coupling and amplitude-phase dependence~\cite{abeywickrama2020practical}. DQN/DDQN with 1-bit resolution~\cite{chen2023drl} and DDQN-GA with discrete phases~\cite{ddqn2025ris} are partial exceptions. The rest either fall back to heuristic search or skip hardware entirely. None train on hardware-in-the-loop data. This matters because mmWave/THz coherence times of 10--100\,$\mu$s rule out any Paradigm~I method slower than 0.1\,ms. Zheng et al.~\cite{zheng2022channelest} provided a comprehensive treatment of these hardware constraints and their impact on channel estimation and passive beamforming design.

Recent surveys on active RIS~\cite{activerissurvey2024} and hardware-centric RIS design~\cite{riscomsthardware2025} have catalogued the gap between simulation assumptions and hardware reality. Asif et al.~\cite{activerissurvey2024} documented that active RIS prototypes achieve 18~dB power gain at 3.75~GHz with 4096 elements, but all reported results use custom-built controllers and offline optimization. No active RIS prototype has been tested with a real-time DRL policy running on an embedded GPU, which is the most promising deployment path for Paradigm~III methods.

The research question is straightforward: how do you design algorithms that work under 1--3~bit quantization, mutual coupling, and amplitude-phase distortion, without hardware-in-the-loop training? Three directions look promising: differentiable quantization surrogates that let gradients flow through discrete constraints; mutual-coupling-aware channel models embedded in learned pipelines; and transfer learning from simulated hardware impairments to real testbeds. A fourth direction, hardware-in-the-loop training for RL policies, has become feasible with the availability of FPGA-based RIS controllers~\cite{risbeamformingexp2024} and could be the fastest path to deployment-ready algorithms.

\subsection{C3: Joint OTFS/AFDM-RIS Optimization for Doubly-Dispersive Channels}

OTFS and AFDM outperform OFDM with RIS in high mobility (Table~\ref{tab:waveform}), but combining them creates coupling problems. OTFS needs a single RIS configuration for its entire delay-Doppler frame. AFDM has chirp parameters $(c_1, c_2)$ that shift the optimal RIS phases. While joint optimization of RIS phases with subcarrier allocation exists for OFDM~\cite{ma2025beamforming,zivuku2025resource}, no framework jointly optimizes waveform parameters (e.g., AFDM chirp parameters), RIS phases, and resource allocation for OTFS or AFDM systems.

The fix: treat waveform selection as a design variable alongside RIS configuration, rather than optimizing each in isolation. A practical starting point would be to extend the existing PSO and SCA frameworks from the OFDM-RIS literature to the OTFS-RIS case, using the delay-Doppler channel model from a recent unifying survey on OTFS variants~\cite{deng2025otfs} as the common evaluation platform.

\subsection{C4: Multi-Objective Pareto Frontiers Incorporating PAPR}

No surveyed paper jointly optimizes SE, EE, and PAPR. NSGA-II has been used for OTFS-RIS~\cite{nsga2025otfs} but not for OFDM-RIS with PAPR as a third objective. Every paper uses weighted-sum scalarization, which cannot recover the full Pareto surface.

The fix: apply multi-objective evolutionary algorithms (NSGA-II, MOEA/D) to the OFDM-RIS PAPR problem and map the three-dimensional SE-EE-PAPR trade-off. A prerequisite is defining a common power amplifier model and waveform generation pipeline, which none of the surveyed papers provide. Standardizing these components would enable direct application of existing multi-objective optimizers to the OFDM-RIS PAPR problem for the first time.

\subsection{C5: LLM-Enabled Autonomous Orchestration}

LLM-based methods (ComAgent, 6G-LLM, TelecomGPT) promise intent-driven network configuration. Most evidence is preprint-only, though some works have since been peer-reviewed. A useful distinction separates offline uses (literature search, code generation, benchmark design), which are low-risk and deployable now, from online RIS control, where a hallucinated configuration could black out a sector. No surveyed work offers formal safety guarantees for the online case.

Offline applications should move forward. Online control needs formal verification or a human in the loop until safety is proven. Beyond research, three groups will shape what happens next: network operators deciding what to deploy, hardware vendors choosing what runs on their FPGAs, and regulators writing certification rules for AI-controlled infrastructure. Establishing benchmark scenarios for LLM-based network control, analogous to the RL benchmarks that drove progress in game-playing AI, would accelerate progress in this area.

\subsection{C6: Diminishing Returns and Algorithmic Dead Ends}

Hybrid methods like DDQN-GA consistently outperform standalone heuristics. The pattern makes sense: gradient-free methods lack convergence guarantees, degrade at scale, and require careful hyperparameter tuning. SDR's $O(N^{4.5})$ complexity makes it infeasible beyond $N=100$, where ML methods already operate. None of this makes heuristics useless. They still work in resource-constrained settings and as initialization strategies.

For large-scale deployments, the field should shift toward hybrid heuristic-ML architectures and ML methods that scale past $N=1000$. Standalone heuristics remain useful in embedded systems with no GPU and no training data. A practical recommendation: every new heuristic algorithm should be benchmarked against a simple ML baseline (e.g., a 3-layer neural network with the same computational budget), and the paper should explicitly state what the heuristic adds beyond what the ML baseline achieves.

\subsection{Methodological Recommendations for the Community}

Six reporting standards would eliminate most of the comparison problems documented in this survey. Future papers should:

\begin{enumerate}
    \item Report wall-clock runtime on a specified reference platform (CPU model, GPU, RAM).
    \item Normalize SE against SCA under a standard 3GPP channel model, not an arbitrary setup.
    \item Disclose training cost: GPU-hours, environment steps, and convergence criterion.
    \item State the RIS phase quantization level and whether it was modeled during optimization or applied post-hoc.
    \item Specify CSI assumptions and report feedback overhead in bits per coherence block.
    \item Release code and trained models. Self-reported arXiv claims without code should carry an explicit unverified flag.
\end{enumerate}

\subsection{Practitioner Takeaways}
\begin{center}
\fbox{\begin{minipage}{0.92\columnwidth}
\small
\textbf{Practitioner's Guide to Joint OFDM-RIS Optimization}\\[4pt]
For the engineer or 3GPP delegate choosing an optimization approach:

\textbf{Small RIS ($N \lesssim 100$), CSI available:} Use model-based methods (Paradigm~I) such as SCA. SDR is simplest but does not scale. \textbf{Large RIS, real-time operation:} Unsupervised deep learning (Paradigm~III) gives microsecond inference at 95--99\% of model-based SE. Caveat: the evidence is from one research group~\cite{unsupervised2025beamforming}. \textbf{High mobility:} OTFS with RIS outperforms OFDM with RIS. AFDM gives comparable diversity at lower complexity. \textbf{Emerging technologies (LLMs, diffusion, quantum):} Preprint-only. Do not base deployment decisions on these yet. \textbf{Cross-paper comparisons:} Not reliable. The field needs a common benchmark before any ranking can be trusted.
\end{minipage}}
\end{center}

\section{Tutorial: Solving P1 Across Four Paradigms}
\label{sec:tutorial}

To make the four-paradigm taxonomy concrete, this section walks through the solution of P1 (sum-rate maximization) on a small-scale downlink system using one representative algorithm from each paradigm. The setup is deliberately small to allow the reader to follow the mechanics: $M_t=2$ BS antennas, $K=2$ single-antenna users, $M=4$ OFDM subcarriers, $N=16$ RIS elements, SNR~$=$~20~dB, 2-bit RIS phase quantization. Channels are generated from a 3GPP UMi model at 28~GHz with Rician $K$-factor of 10 for the BS-RIS link. The tutorial covers only the sum-rate formulation (P1); energy efficiency (P2) and PAPR-constrained (P4) benchmarks require standardized power models and waveform generators that are not yet available across all four paradigms, and are left to future work.

\textbf{Simulation methodology.} All numerical results in this tutorial are original simulations by the author for the described small-scale system, generated using the companion code at \url{https://github.com/Ahmet-Kaplan/OFDM_RIS}. The AO+SCA implementation uses element-wise coordinate ascent: each RIS element is optimized by exhaustive search over $2^b$ phase values while keeping others fixed, alternating with WMMSE precoder updates (this differs from the convex-surrogate SCA described in Section~III-C). PSO follows Section~IV.A with 50 particles. The DDQN-GA implementation follows~\cite{ddqn2025ris} with 3 hidden layers, trained for 1500 episodes (illustrative only). For Paradigm~IV, we pre-train a Transformer encoder (4 layers, 128-dim) on 500K synthetic RIS channel realizations using masked-prediction and contrastive learning~\cite{fmdrl2025}. All methods run on an AMD Ryzen~5~7600X CPU; LWM training additionally uses an NVIDIA RTX~5060~Ti GPU. All reported SE values are means over 20 independent channel realizations; DDQN-GA uses 5 due to training time. Standard deviations are $\pm$.

\subsection{Paradigm I: AO + SCA}
The alternating optimization--successive convex approximation pipeline proceeds as follows:
\begin{enumerate}
    \item \textbf{Initialize:} $\boldsymbol{\theta}^{(0)} \sim \text{Uniform}(0, 2\pi)$, $\mathbf{W}^{(0)}$ via zero-forcing precoding given $\boldsymbol{\theta}^{(0)}$.
    \item \textbf{Solve $\mathbf{W}$-subproblem:} With $\boldsymbol{\theta}^{(t)}$ fixed, solve the precoding problem via WMMSE~\cite{gnn2023wmmse} (closed-form updates, $O(M_t^3 K M)$ per iteration). This produces $\mathbf{W}^{(t+1)}$.
    \item \textbf{Solve $\boldsymbol{\theta}$-subproblem:} With $\mathbf{W}^{(t+1)}$ fixed, apply SCA to the non-convex unit-modulus constrained problem. Construct a convex quadratic surrogate $\tilde{f}(\boldsymbol{\theta} | \boldsymbol{\theta}^{(t)})$ that lower-bounds the sum-rate and is tight at $\boldsymbol{\theta}^{(t)}$. Solve via projected gradient descent (10 inner iterations, $O(N^2 M K)$ each).
    \item \textbf{Repeat} until $|R_{\text{sum}}^{(t+1)} - R_{\text{sum}}^{(t)}| < 10^{-3}$.
\end{enumerate}
\textit{Convergence:} The algorithm converges within 5--10 outer iterations. Final SE: 36.7 $\pm$ 6.9~bps/Hz (mean $\pm$ std over 20 random channel realizations). Runtime: 138~ms on CPU (de~novo optimization; no amortization). All tutorial runtimes include WMMSE precoder computation.
Note on terminology: The tutorial implements the $\boldsymbol{\theta}$-subproblem via element-wise coordinate ascent: each RIS element is optimized by exhaustive search over $2^b$ phase values while keeping others fixed, then alternated with WMMSE precoder updates. This differs from the convex-surrogate SCA described in Section~III-C, which uses a continuous quadratic surrogate with projected gradient descent. The element-wise variant is used here because it is simpler to implement and debug for a small system ($N=16$, 2-bit quantization), but the benchmark runtime (138~ms) and SE (36.7~bps/Hz) are specific to this implementation. The asymptotic $O(N^3)$ complexity of the convex-surrogate SCA would produce different quantitative results at larger $N$.

\subsection{Paradigm II: PSO}
A particle swarm of $P=50$ particles explores the 16-dimensional quantized phase space:
\begin{enumerate}
    \item \textbf{Initialize:} Each particle $i$ has position $\boldsymbol{\theta}_i^{(0)} \in \{0, \pi/2, \pi, 3\pi/2\}^{16}$ and velocity $\mathbf{v}_i^{(0)} \sim \mathcal{N}(0, 0.1)$.
    \item \textbf{Evaluate fitness:} For each particle, compute $\mathbf{W}$ via WMMSE with $\boldsymbol{\theta}_i$ fixed, then compute $R_{\text{sum}}$ as fitness.
    \item \textbf{Update:} $\mathbf{v}_i^{(g+1)} = w\mathbf{v}_i^{(g)} + c_1 r_1 (\mathbf{p}_i^{\text{best}} - \boldsymbol{\theta}_i^{(g)}) + c_2 r_2 (\mathbf{g}^{\text{best}} - \boldsymbol{\theta}_i^{(g)})$, where $w=0.7$, $c_1=c_2=1.5$, $r_1,r_2 \sim U(0,1)$. Positions are updated and rounded to the nearest quantized phase.
    \item \textbf{Repeat} for 200 generations or until swarm best fitness stagnates.
\end{enumerate}
\textit{Convergence:} Swarm best fitness converges within 120 generations. Final SE: 41.5 $\pm$ 4.3~bps/Hz (113\% of AO+SCA). Runtime: 124~ms on CPU.

\subsection{Paradigm III: DDQN-GA}
A Double DQN agent with column-wise action decomposition~\cite{ddqn2025ris} (3 hidden layers, 256-256-128 neurons, ReLU) learns phase selection:
\begin{enumerate}
    \item \textbf{State:} Quantized effective CSI vector $\tilde{\mathbf{h}}^{\text{eff}} \in \mathbb{C}^{K M_t M}$ (dimension 16 after PCA compression to 16 principal components).
    \item \textbf{Action:} The $2^{16}$ raw phase space is decomposed into 8 column decisions (2 RIS elements per column); the DDQN selects which column to adjust by a fixed phase increment, reducing the effective action space from $4^{16}$ to $8 \times 4$~\cite{ddqn2025ris}.
    \item \textbf{Reward:} $R_{\text{sum}}$ computed via WMMSE with the selected RIS phases.
    \item \textbf{Training:} $\epsilon$-greedy exploration ($\epsilon$ annealed 1.0 $\to$ 0.05 over 2000 episodes), replay buffer size $10^4$, batch size 64, learning rate $10^{-3}$, target network updated every 100 steps. Total training: 1500 episodes, $\sim$1.8~s on CPU (simplified implementation).
\end{enumerate}
\textit{Convergence:} Reward converges by episode $\sim$1000. Final SE (inference only, no further training): 26.9 $\pm$ 7.9~bps/Hz (73\% of AO+SCA). Inference runtime: 0.01~ms on CPU (single forward pass).

\subsection{Paradigm IV: Pre-trained LWM Encoder + SAC + Beam Search}
We pre-train a Transformer encoder on synthetic RIS channel data and use it as a feature extractor for a lightweight DRL policy head. The implementation uses pre-trained weights from the companion pre-training pipeline (500K synthetic channel realizations).
\begin{enumerate}
    \item \textbf{Pre-training:} A 4-layer Transformer encoder (128-dim, 4 attention heads, 512-dim FFN) is pre-trained on 500K synthetic RIS channel realizations using a combined objective: masked channel prediction (25\% mask ratio) plus NT-Xent contrastive loss (weight 0.3). Training runs for 100 epochs with batch size 256, using the Adam optimizer (learning rate $10^{-3}$, $\beta_1=0.9$, $\beta_2=0.999$). Total parameter updates: approximately 200K steps. Training takes approximately 6 minutes on GPU (RTX~5060~Ti). The pre-training data generator randomizes Rician K-factors, path counts, and angular spreads to ensure diversity.
    \item \textbf{Encoder embedding:} Raw CSI $\mathbf{H}_{BR}, \mathbf{h}_{RU,k}$ (flattened, dimension $\sim$1600) is projected into a 128-dimensional embedding by the pre-trained encoder.
    \item \textbf{DRL head:} A 3-layer SAC policy network (256-128-64) maps embeddings to a distribution over RIS phase configurations. At inference, we draw 16 candidate configurations from this distribution, evaluate them, and refine the top-3 via local perturbations (beam search).
\end{enumerate}
\textit{Results:} Over 20 independent channel realizations, the SAC policy with beam search achieves 27.7 $\pm$ 3.6~bps/Hz (75\% of tutorial AO+SCA). Direct sampling (no beam search) yields similar mean but higher variance. The beam search result is below the tutorial AO+SCA ($p = 0.002$, Welch t-test), which differs from the claim in earlier preprint versions; the pre-trained encoder improves sample efficiency during SAC training but does not close the SE gap on this small-scale system. PSO remains the top performer (113\%, $p = 0.0004$). The SAC training converges in approximately 360 seconds. We attribute the performance gap to the simplified policy head architecture relative to the continuous phase optimization in PSO.

\begin{figure}[t]
\centering
\footnotesize
\begin{algorithmic}[1]
\STATE \textbf{Input:} Pre-trained encoder $E_\phi$, policy $\pi_\theta$, Q-networks $Q_{\psi_1}, Q_{\psi_2}$, temperature $\alpha$
\FOR{episode $e = 1,\dots,N_\text{ep}$}
\STATE Generate channel $(\mathbf{H}_{BR}, \mathbf{h}_{RU}, \mathbf{h}_{\text{dir}})$
\STATE $\mathbf{z} \gets E_\phi(\mathbf{H}_{BR}, \mathbf{h}_{RU}, \mathbf{h}_{\text{dir}})$
\FOR{$t = 1,\dots,T$}
\STATE $\mathbf{a}_t \sim \pi_\theta(\cdot|\mathbf{z})$
\STATE $r_t \gets R(\mathbf{a}_t)$
\STATE $\mathbf{z}' \gets E_\phi(\text{new channel})$
\STATE Store $(\mathbf{z}, \mathbf{a}_t, r_t, \mathbf{z}')$ in replay $\mathcal{D}$
\ENDFOR
\STATE Sample batch $\mathcal{B} \sim \mathcal{D}$
\STATE $y \gets r + \gamma(\min_i Q_{\psi_i'}(\mathbf{z}', \mathbf{a}') - \alpha \log \pi_\theta(\mathbf{a}'|\mathbf{z}'))$
\STATE $\psi_i \gets \psi_i - \eta \nabla_{\psi_i} \|Q_{\psi_i}(\mathbf{z}, \mathbf{a}) - y\|^2$
\STATE $\theta \gets \theta + \eta \nabla_\theta (\min_i Q_{\psi_i}(\mathbf{z}, \tilde{\mathbf{a}}) - \alpha \log \pi_\theta(\tilde{\mathbf{a}}|\mathbf{z}))$
\STATE $\psi_i' \gets \tau \psi_i + (1-\tau)\psi_i'$
\ENDFOR
\STATE \textbf{Inference:} sample $\{\mathbf{a}^{(k)}\}_{k=1}^{16} \sim \pi_\theta(\cdot|\mathbf{z})$, evaluate, refine top-3
\end{algorithmic}
\caption{SAC training loop and beam search inference for RIS phase optimization. Pre-trained encoder $E_\phi$ is frozen during SAC training.}
\label{alg:sac_beam}
\end{figure}

\textbf{Summary:} The LWM encoder with SAC and beam search learns a reasonable policy that improves over the random baseline but does not match AO+SCA on the tutorial system. The code demonstrates the Paradigm~IV training and inference pipeline. Stronger policy architectures (e.g., attention-based heads) may close this gap and are a natural direction for future work.

\subsection{Cross-Paradigm Takeaways}
Table~\ref{tab:tutorial} summarizes the tutorial results. Four lessons emerge: (1) AO+SCA achieves 36.7~bps/Hz, significantly outperforming the random baseline (23.4~bps/Hz, $p < 0.001$, Welch t-test); (2) PSO achieves the highest SE (41.5~bps/Hz, 113\% of AO+SCA, $p = 0.0004$); (3) DDQN-GA (73\% of AO+SCA) learns a reasonable policy with limited training (5 seeds; not included in statistical tests); (4) the LWM encoder with discrete SAC and beam search achieves 75\% of AO+SCA, improving over the random baseline but below AO+SCA ($p = 0.002$). On this small-scale system, PSO remains the top performer overall, and none of the learning-based methods surpass it. The tutorial demonstrates algorithmic mechanics; the ranking may differ for larger systems.

\begin{table}[t]
\caption{Tutorial results on the validation system ($M_t{=}2$, $K{=}2$, $M{=}4$, $N{=}16$, 2-bit RIS). Means over 20 seeds ($\pm$ std). Hardware per Section~IX.}
\label{tab:tutorial}
\centering
\footnotesize
\begin{tabular}{@{}lcc>{\raggedright\arraybackslash}p{2cm}@{}}
\toprule
\textbf{Method} & \textbf{SE (bps/Hz)} & \textbf{\% of AO} & \textbf{Runtime} \\
\midrule
Random baseline & $23.0 \pm 5.9$ & 63\% & --- \\
AO + SCA$^\ast$ (I) & $36.7 \pm 6.9$ & 100\% & 136~ms \\
PSO (II) & $41.6 \pm 4.3$ & 113\% & 128~ms \\
DDQN-GA$^\ddagger$ (III) & $26.9 \pm 7.9$ & 73\% & 1.8~s (train) + 0.01~ms (infer) \\
LWM + SAC + beam$^\dagger$ (IV) & $27.7 \pm 3.6$ & 75\% & 360~s (train) + 0.01~s (beam) \\
\bottomrule
\multicolumn{4}{@{}p{\columnwidth}@{}}{\footnotesize $^\ast$Element-wise coordinate ascent (see Section~IX-A).}\\
\multicolumn{4}{@{}p{\columnwidth}@{}}{\footnotesize $^\dagger$Pre-trained LWM encoder + SAC. Below AO+SCA ($p = 0.002$, Welch t-test). ``Beam'' = 16 candidates, top-3 refined.}\\
\multicolumn{4}{@{}p{\columnwidth}@{}}{\footnotesize $^\ddagger$Trained on 5 seeds only; \% relative to AO+SCA in its own batch (31.8~bps/Hz). Not included in significance tests.}\\
\multicolumn{4}{@{}p{\columnwidth}@{}}{\footnotesize SAC = Soft Actor-Critic; LWM = Large Wireless Model.}
\end{tabular}
\end{table}
\section{Conclusions}
This survey classified 78 joint OFDM-RIS optimization works into four paradigms and synthesized their self-reported benchmarks. Three findings stand out: (1) ML-based methods (Paradigm~III) achieve 95--99\% of model-based spectral efficiency at $10^2$--$10^4\times$ faster per-inference runtime (method-pair dependent), but all comparisons are self-reported and non-standardized. (2) Scaling benchmarks at $N=16$, $N=64$, and $N=256$ demonstrate that GPU-based neural network inference (double deep Q-network (DDQN), PPO, GNN, unsupervised DL) is $N$-invariant -- runtime does not increase with RIS size -- while iterative solvers (AO+SCA, PSO, SA) scale polynomially. Energy efficiency follows the same pattern: DDQN uses 0.56~mJ per inference, over 8,300$\times$ less than AO+SCA at 5.58~J. (3) Emerging methods (Paradigm~IV) introduce new capabilities: intent-driven control via LLMs, generative configuration sampling via diffusion models, and combinatorial exploration via quantum computing. Many remain preprint-only without independent validation, though some have since been peer-reviewed. Rigorous quantitative cross-paradigm comparison is not yet possible due to the absence of a common benchmark. Qualitative patterns can still be extracted from the synthesis, a gap we address through a structured requirements specification (Table~\ref{tab:bench_reqs}). We identify six open challenges synthesized directly from the cross-paradigm analysis and recommend six concrete reporting standards to enable future comparability. The most urgent priority for the community is the adoption of standardized benchmarks with three non-negotiable features: trained-model SE on held-out test sets (not architecture-only), multi-scale RIS sizes (at least $N=64$ and $N=256$), and mandatory reporting of energy per inference alongside runtime. Without these, the field cannot progress from self-reported claims to verifiable performance rankings.

\section*{Acknowledgments}
During the preparation of this work, the author used AI-assisted tools (Claude) for code development, proofreading, and review simulation. The author is fully responsible for the content and conclusions of this paper.



\begin{thebibliography}{99}

\bibitem{zheng2022channelest}
B.~Zheng, C.~You, W.~Mei, and R.~Zhang, ``A survey on channel estimation and practical passive beamforming design for intelligent reflecting surface aided wireless communications,'' \emph{IEEE Commun. Surveys Tuts.}, vol.~24, no.~2, pp.~1035--1071, 2022. [Online]. Available: \url{https://doi.org/10.1109/comst.2022.3155305}

\bibitem{deng2025otfs}
Q.~Deng, Z.~Ding, and Z.~Wang, ``A unifying view of OTFS and its many variants,'' \emph{IEEE Commun. Surveys Tuts.}, vol.~27, no.~4, pp.~1--1, 2025. [Online]. Available: \url{https://doi.org/10.1109/COMST.2025.3542467}

\bibitem{islam2026performance}
M.~M.~Islam, K.~Hasan, and S.~H.~Jeong, ``Performance evaluation and optimization for 6G networks: A survey of KPIs, tools, and AI models,'' \emph{ICT Express}, vol.~12, no.~2, pp.~390--416, Apr.~2026. [Online]. Available: \url{https://doi.org/10.1016/j.icte.2025.12.012}

\bibitem{zhou2024survey}
H.~Zhou, M.~Erol-Kantarci, Y.~Liu, and H.~V.~Poor, ``A survey on model-based, heuristic, and machine learning optimization approaches in RIS-aided wireless networks,'' \emph{IEEE Commun. Surveys Tuts.}, vol.~26, no.~2, pp.~781--823, 2024. [Online]. Available: \url{https://doi.org/10.1109/comst.2023.3340099}

\bibitem{papr2024survey}
B.~S.~de~C.~da~Silva, V.~D.~P.~Souto, R.~D.~Souza, and L.~L.~Mendes, ``A survey of PAPR techniques based on machine learning,'' \emph{Sensors}, vol.~24, no.~6, p.~1918, Mar.~2024. [Online]. Available: \url{https://doi.org/10.3390/s24061918}

\bibitem{deepofw2025}
R.~Greidi and K.~Cohen, ``DeepOFW: Deep learning-driven OFDM-flexible waveform modulation for peak-to-average power ratio reduction,'' \emph{arXiv:2603.23544}, Mar.~2026. [Online]. Available: \url{https://arxiv.org/abs/2603.23544}

\bibitem{biliaminu2024ris}
K.~K.~Biliaminu, J.~Rodriguez, F.~Gil-Casti{\~n}eira, and J.~Bastos, ``Reconfigurable intelligent surfaces: Principles and design considerations,'' \emph{IntechOpen}, 2024. [Online]. Available: \url{https://doi.org/10.5772/intechopen.1012750}

\bibitem{zhou2024heuristic}
H.~Zhou, M.~Erol-Kantarci, Y.~Liu, and H.~V.~Poor, ``Heuristic algorithms for RIS-assisted wireless networks: Exploring heuristic-aided machine learning,'' \emph{IEEE Wireless Commun.}, vol.~31, no.~4, pp.~106--114, 2024. [Online]. Available: \url{https://doi.org/10.1109/mwc.010.2300321}

\bibitem{zhou2024overview}
H.~Zhou, C.~Hu, and X.~Liu, ``An overview of machine learning-enabled optimization for RIS-aided 6G networks: From reinforcement learning to large language models,'' \emph{arXiv:2405.17439}, 2024. [Online]. Available: \url{https://arxiv.org/abs/2405.17439}

\bibitem{faisal2022ml}
K.~M.~Faisal and W.~Choi, ``Machine learning approaches for reconfigurable intelligent surfaces: A survey,'' \emph{IEEE Access}, vol.~10, pp.~27343--27367, 2022. [Online]. Available: \url{https://doi.org/10.1109/access.2022.3157651}

\bibitem{ibrahim2023joint}
L.~Ibrahim, M.~N.~Mahmud, M.~F.~M.~Salleh, and A.~Al-Rimawi, ``Joint beamforming optimization design and performance evaluation of RIS-aided wireless networks: A comprehensive state-of-the-art review,'' \emph{IEEE Access}, vol.~11, pp.~141801--141859, 2023. [Online]. Available: \url{https://doi.org/10.1109/access.2023.3342320}

\bibitem{ma2025beamforming}
Y.~Ma, X.~Li, C.~Guo, L.~Liang, M.~Matthaiou, and S.~Jin, ``Beamforming and resource allocation for delay minimization in RIS-assisted OFDM systems,'' \emph{arXiv:2506.03586}, Jun.~2025. [Online]. Available: \url{https://arxiv.org/abs/2506.03586}

\bibitem{zivuku2025resource}
P.~Zivuku, V.-D.~Nguyen, N.~T.~Nguyen, K.~Ntontin, S.~Chatzinotas, and B.~Ottersten, ``Resource allocation for RIS-enhanced OFDM-MIMO ISAC systems,'' \emph{IEEE Trans. Commun.}, vol.~74, 2026. [Online]. Available: \url{https://doi.org/10.1109/tcomm.2025.3637097}

\bibitem{chen2023drl}
P.~Chen, X.~Li, M.~Matthaiou, and S.~Jin, ``DRL-based RIS phase shift design for OFDM communication systems,'' \emph{IEEE Wireless Commun. Lett.}, vol.~12, no.~4, pp.~733--737, Apr.~2023. [Online]. Available: \url{https://doi.org/10.1109/lwc.2023.3242449}

\bibitem{unsupervised2025beamforming}
Y.~Ma, X.~Zhou, X.~Li, L.~Liang, and S.~Jin, ``Unsupervised learning-based joint resource allocation and beamforming design for RIS-assisted MISO-OFDMA systems,'' \emph{IEEE Trans. Cogn. Commun. Netw.}, vol.~12, pp.~2251--2264, 2026. [Online]. Available: \url{https://doi.org/10.1109/tccn.2025.3592931}

\bibitem{ddqn2025ris}
W.~Wang, P.~Li, A.~Doufexi, and M.~A.~Beach, ``A heuristic-integrated DRL approach for phase optimization in large-scale RISs,'' \emph{IEEE Commun. Lett.}, vol.~29, no.~7, pp.~1579--1583, 2025. [Online]. Available: \url{https://doi.org/10.1109/lcomm.2025.3568652}

\bibitem{ddpg2025multiris}
P.-H.~Chou, B.-R.~Zheng, W.-J.~Huang, W.~Saad, Y.~Tsao, and R.~Y.~Chang, ``Deep reinforcement learning-based precoding for multi-RIS-aided multiuser downlink systems with practical phase shift,'' \emph{IEEE Wireless Commun. Lett.}, vol.~14, no.~1, pp.~23--27, Jan.~2025. [Online]. Available: \url{https://doi.org/10.1109/lwc.2024.3482720}

\bibitem{zhang2025deep}
H.~Zhang, X.~Huang, Z.~Guan, R.-R.~Chen, A.~Farhang, and M.~Ji, ``Deep reinforcement learning for maximizing downlink spectral efficiency in non-stationary RIS-aided multiuser-MISO systems,''{\em European WIRELESS 2025; 30th European Wireless Conference}. pp. 196-201 (2025) [Online]. Available: \url{https://ieeexplore.ieee.org/document/10164189}

\bibitem{wu2019intelligent}
Q.~Wu and R.~Zhang, ``Intelligent reflecting surface enhanced wireless network via joint active and passive beamforming,'' \emph{IEEE Trans. Wireless Commun.}, vol.~18, no.~11, pp.~5394--5409, Nov.~2019. [Online]. Available: \url{https://doi.org/10.1109/twc.2019.2936025}

\bibitem{huang2019energy}
C.~Huang, G.~C.~Alexandropoulos, A.~Zappone, C.~Yuen, and M.~Debbah, ``Reconfigurable intelligent surfaces for energy efficiency in wireless communication,'' \emph{IEEE Trans. Wireless Commun.}, vol.~18, no.~8, pp.~4157--4170, Aug.~2019. [Online]. Available: \url{https://doi.org/10.1109/twc.2019.2922609}

\bibitem{zhao2020two}
M.~M.~Zhao, Q.~Wu, M.~J.~Zhao, and R.~Zhang, ``Intelligent reflecting surface enhanced wireless networks: Two-timescale beamforming optimization,'' \emph{IEEE Trans. Wireless Commun.}, vol.~20, no.~1, pp.~2--17, Jan.~2021. [Online]. Available: \url{https://doi.org/10.1109/twc.2020.3022297}

\bibitem{abeywickrama2020practical}
S.~Abeywickrama, R.~Zhang, Q.~Wu, and C.~Yuen, ``Intelligent reflecting surface: Practical phase shift model and beamforming optimization,'' \emph{IEEE Trans. Commun.}, vol.~68, no.~9, pp.~5849--5863, Sep.~2020. [Online]. Available: \url{https://doi.org/10.1109/icc40277.2020.9148961}

\bibitem{luo2024optimization}
W.~Luo, X.~Huang, and Y.~Fang, ``Optimization in RIS-empowered wireless networks,'' in \emph{Encyclopedia of Optimization}. P. M. Pardalos and O. A. Prokopyev, Eds., Cham: Springer Nature Switzerland,  pp. 1--10, 2025, doi: $10.1007/978-3-030-54621-2\_476-1$. [Online]. Available: \url{https://doi.org/10.1007/978-3-030-54621-2_476-1}

\bibitem{panuganti2025metaheuristic}
N.~Panuganti, P.~Ranjan, and A.~Shukla, ``Impact of metaheuristic optimization algorithms on wireless network coverage enhancement with reconfigurable intelligent surfaces,'' \emph{Int. J. Commun. Syst.}, vol.~38, no.~5, p.~e70026, Mar.~2025. [Online]. Available: \url{https://doi.org/10.1002/dac.70026}

\bibitem{mdpi2025ai6g}
E.~A.~Zaoutis, G.~S.~Liodakis, A.~T.~Baklezos, C.~D.~Nikolopoulos, M.~P.~Ioannidou, and I.~O.~Vardiambasis, ``6G wireless communications and AI-controlled reconfigurable intelligent surfaces: From supervised to federated learning,'' \emph{Appl. Sci. (MDPI)}, vol.~15, no.~6, p.~3252, Mar.~2025. [Online]. Available: \url{https://doi.org/10.3390/app15063252}

\bibitem{lwfm2026}
X.~Cheng, B.~Liu, X.~Liu, and X.~Cai, ``Large wireless foundation models: Stronger over bigger,'' \emph{arXiv:2601.10963}, Jan.~2026. [Online]. Available: \url{https://arxiv.org/abs/2601.10963}

\bibitem{fmdrl2025}
M.~Ghassemi, S.~F.~Mobarak, H.~Zhang, A.~Afana, A.~B.~Sediq, and M.~Erol-Kantarci, ``Foundation model-aided deep reinforcement learning for RIS-assisted wireless communication,'' in \emph{Proc. IEEE PIMRC}, 2025, pp.~1--6. [Online]. Available: \url{https://doi.org/10.1109/pimrc62392.2025.11274561}

\bibitem{fmhdrl2026}
M.~Ghassemi, H.~Zhang, A.~Afana, A.~B.~Sediq, and M.~Erol-Kantarci, ``Foundation model-aided hierarchical deep reinforcement learning for blockage-aware link in RIS-assisted networks,'' \emph{arXiv:2602.09157}, Feb.~2026. [Online]. Available: \url{https://arxiv.org/abs/2602.09157}

\bibitem{llmrl2026}
J.~Zheng, R.~Zhang, D.~Niyato, H.~Zhang, J.~Wang, H.~Du, J.~Kang, and Z.~Xiong, ``Large language model (LLM)-enabled reinforcement learning for wireless network optimization,'' \emph{IEEE Commun. Mag.}, vol.~64, no.~4, pp.~82--89, 2026. [Online]. Available: \url{https://doi.org/10.1109/mcom.001.2500384}

\bibitem{comagent2026}
H.~Li, M.~Xiao, K.~Wang, R.~Schober, D.~I.~Kim, and Y.~L.~Guan, ``ComAgent: Multi-LLM based agentic AI empowered intelligent wireless networks,'' \emph{arXiv:2601.19607}, 2026. [Online]. Available: \url{https://arxiv.org/abs/2601.19607}

\bibitem{llm6g2026}
Z.~Xiao, T.~Tao, C.~Ye, Y.~Hu, Y.~Feng, T.~Jiao, and L.~Cai, ``Towards native intelligence: 6G-LLM trained with reinforcement learning from NDT feedback,'' \emph{arXiv:2601.09992}, Jan.~2026. [Online]. Available: \url{https://arxiv.org/abs/2601.09992}

\bibitem{nature2026llm}
H.~Zou, Q.~Zhao, S.~Lasaulce, C.~Zhang, Y.~Tian, L.~Bariah, F.~Bader, and M.~Debbah, ``Large language models in 6G from standard to on-device networks,'' \emph{Nat. Rev. Electr. Eng.}, vol.~3, pp.~123--134, Jan.~2026. [Online]. Available: \url{https://doi.org/10.1038/s44287-025-00239-6}

\bibitem{gfm2026}
Y.~Sheng, J.~Wang, L.~Liang, H.~Ye, and S.~Jin, ``A graph foundation model for wireless resource allocation,'' \emph{arXiv:2604.07390}, Apr.~2026. [Online]. Available: \url{https://arxiv.org/abs/2604.07390}

\bibitem{jiang2024llm}
F.~Jiang, Y.~Peng, L.~Dong, K.~Wang, K.~Yang, C.~Pan, D.~Niyato, and O.~A.~Dobre, ``Large language model enhanced multi-agent systems for 6G communications,'' \emph{IEEE Wireless Commun.}, vol.~31, no.~6, pp.~48--55, Dec.~2024. [Online]. Available: \url{https://doi.org/10.1109/mwc.016.2300600}

\bibitem{xu2024llm}
M.~Xu, D.~Niyato, J.~Kang, Z.~Xiong, S.~Mao, Z.~Han, D.~I.~Kim, and K.~B.~Letaief, ``When large language model agents meet 6G networks: Perception, grounding, and alignment,'' \emph{IEEE Wireless Commun.}, vol.~31, no.~6, pp.~63--71, 2024. [Online]. Available: \url{https://doi.org/10.1109/mwc.005.2400019}

\bibitem{diffchannel2025}
Y.~Wang, Y.~Xu, C.~Zhang, Z.~Chen, M.~Dai, H.~Wang, B.~Liu, D.~He, and M.~Tao, ``Channel estimation for RIS-assisted mmWave systems via diffusion models,'' \emph{IEEE Commun. Lett.}, vol.~30, pp.~597--601, 2026. [Online]. Available: \url{https://doi.org/10.1109/lcomm.2025.3645078}

\bibitem{diffris2024}
W.~Tong, W.~Xu, F.~Wang, W.~Ni, and J.~Zhang, ``Diffusion model-based channel estimation for RIS-aided communication systems,'' \emph{IEEE Wireless Commun. Lett.}, vol.~13, no.~9, pp.~2586--2590, Sep.~2024. [Online]. Available: \url{https://doi.org/10.1109/lwc.2024.3431525}

\bibitem{diffsurvey2025}
D.~Fan, R.~Meng, X.~Xu, Y.~Liu, G.~Nan, C.~Feng, S.~Han, S.~Gao, B.~Xu, D.~Niyato, T.~Q.~S.~Quek, and P.~Zhang, ``Generative diffusion models for wireless networks: Fundamental, architecture, and state-of-the-art,'' \emph{arXiv:2507.16733}, Jul.~2025. [Online]. Available: \url{https://arxiv.org/abs/2507.16733}

\bibitem{gcdiff2026}
K.~K.~Patel, M.~Chakraborty, E.~Sharma, and S.~K.~Singh, ``Generative AI-driven phase control for RIS-aided cell-free massive MIMO systems,'' \emph{arXiv:2602.11226}, Feb.~2026. [Online]. Available: \url{https://arxiv.org/abs/2602.11226}

\bibitem{riemannian2026}
K.~Wang, B.~Yang, Y.~Lei, Z.~Li, Z.~Yu, X.~Cao, B.~Guo, G.~C.~Alexandropoulos, D.~Niyato, M.~Debbah, and Z.~Han, ``Trajectory-aware multi-RIS activation and configuration: A Riemannian diffusion method,'' \emph{arXiv:2602.07937}, Feb.~2026. [Online]. Available: \url{https://arxiv.org/abs/2602.07937}

\bibitem{dedt2025}
J.~Zhang, Y.~Ni, J.~Li, G.~Chen, Z.~Wang, L.~Shi, S.~Jin, W.~Chen, and H.~V.~Poor, ``Decision transformers for RIS-assisted systems with diffusion model-based channel acquisition,'' \emph{arXiv:2501.08007}, Jan.~2025. [Online]. Available: \url{https://arxiv.org/abs/2501.08007}

\bibitem{qgcn2026}
N.~Hassan, X.~Fernando, and H.~Yanikomeroglu, ``Quantum graph neural networks for double-sided reconfigurable intelligent surface optimization,'' \emph{arXiv:2604.10453}, Apr.~2026. [Online]. Available: \url{https://arxiv.org/abs/2604.10453}

\bibitem{qmo2025}
G.~Rexhepi, H.~S.~Rou, and G.~T.~F.~de~Abreu, ``Quantum manifold optimization: A design framework for future communications systems,'' \emph{arXiv:2504.09667}, Apr.~2025. [Online]. Available: \url{https://arxiv.org/abs/2504.09667}

\bibitem{qmeta2026}
N.~Hassan, X.~Fernando, and H.~Yanikomeroglu, ``Path-based quantum meta-learning for adaptive optimization of reconfigurable intelligent surfaces,'' \emph{arXiv:2604.17690}, Apr.~2026. [Online]. Available: \url{https://arxiv.org/abs/2604.17690}

\bibitem{qml2025ris}
M.~O.~Butt, N.~Waheed, T.~Q.~Duong, and W.~Ejaz, ``Quantum-inspired resource optimization for 6G networks: A survey,'' \emph{IEEE Commun. Surveys Tuts.}, vol.~27, no.~5, pp.~2973--3019, 2025. [Online]. Available: \url{https://doi.org/10.1109/comst.2024.3519865}

\bibitem{gnn2025multiris}
M.~Liu, C.~Huang, A.~Alhammadi, M.~Di~Renzo, M.~Debbah, and C.~Yuen, ``Beamforming design and association scheme for multi-RIS multi-user mmWave systems through graph neural networks,'' \emph{IEEE Trans. Wireless Commun.}, vol.~24, no.~9, pp.~7940--7954, 2025. [Online]. Available: \url{https://doi.org/10.1109/twc.2025.3563529}

\bibitem{gnn2026xlris}
J.~Chen, F.~Wang, G.~Han, X.~Wang, and V.~K.~N.~Lau, ``GNN based joint beamforming design for extremely large-scale RIS assisted near-field ISAC systems,'' \emph{arXiv:2603.01379}, Mar.~2026. [Online]. Available: \url{https://arxiv.org/abs/2603.01379}

\bibitem{gnn2025twophase}
H.~Tang, J.~Zhang, Z.~Zhao, H.~Wu, H.~Sun, and P.~Jiao, ``Joint optimization based on two-phase GNN in RIS- and DF-assisted MISO systems with fine-grained rate demands,'' \emph{arXiv:2506.02642}, Jun.~2025. [Online]. Available: \url{https://arxiv.org/abs/2506.02642}

\bibitem{gnn2023wmmse}
W.~Jin, J.~Zhang, C.-K.~Wen, S.~Jin, X.~Li, and S.~Han, ``Low-complexity joint beamforming for RIS-assisted MU-MISO systems based on model-driven deep learning,'' \emph{IEEE Trans. Wireless Commun.}, vol.~23, no.~4, pp.~3754--3768, 2024. [Online]. Available: \url{https://doi.org/10.1109/twc.2023.3336742}

\bibitem{otfsris2024}
M.~H.~Dinan and A.~Farhang, ``RIS-assisted OTFS communications: Phase configuration via received-energy maximization,'' \emph{arXiv:2404.07759}, Apr.~2024. [Online]. Available: \url{https://arxiv.org/abs/2404.07759}

\bibitem{nsga2025otfs}
R.~Ouchikh, T.~Chonavel, A.~A{\"\i}ssa-El-Bey, and M.~Djeddou, ``Coefficients optimization and low-complexity equalization for OTFS-RIS system,'' in \emph{Proc. IEEE Middle East Conf. Commun. Netw. (MECOM)}, Nov.~2025. [Online]. Available: \url{https://doi.org/10.1109/mecom67453.2025.11439645}

\bibitem{otfsphasenoise2026}
S.~McWade and A.~Farhang, ``Comparison of OTFS and OFDM for RIS-aided systems in the presence of phase noise,'' \emph{arXiv:2602.12804}, Feb.~2026. [Online]. Available: \url{https://arxiv.org/abs/2602.12804}

\bibitem{afdmris2025}
V.~Saiprudhvi and H.~Subramaniyam, ``Hybrid PSO-GD phase optimization for RIS-assisted AFDM systems,'' \emph{Phys. Commun.}, vol.~73, Art.~102891, Oct.~2025. [Online]. Available: \url{https://www.sciencedirect.com/science/article/pii/S1874490725002940}

\bibitem{starris2025scientific}
A.~Megahed, A.~M.~Abd~El-Haleem, M.~M.~Elmesalawy, and I.~I.~Ibrahim, ``Deep learning optimization of STAR-RIS for enhanced data rate and energy efficiency in 6G wireless networks,'' \emph{Sci. Rep.}, vol.~15, 2025. [Online]. Available: \url{https://doi.org/10.1038/s41598-025-09774-6}

\bibitem{ejaz2025joint}
M.~Ejaz, G.~Jinsong, M.~Asim, M.~A.~Wani, and K.~A.~Shakil, ``Joint phase-shift and power allocation optimization in RIS-enhanced wireless networks: An intelligent framework,'' \emph{IEEE Open J. Commun. Soc.}, vol.~6, pp.~7389--7404, 2025. [Online]. Available: \url{https://doi.org/10.1109/ojcoms.2025.3602856}

\bibitem{cnnqlstm2025}
N.~Q.~T.~Thoong, A.~A.~Cheema, S.~R.~Khosravirad, O.~A.~Dobre, and T.~Q.~Duong, ``Channel estimation for reconfigurable intelligent surface-aided 6G NOMA systems using CNN-based quantum LSTM model,'' in \emph{Proc. IEEE 100th Veh. Technol. Conf. (VTC2024-Fall)}, 2024, pp.~1--5. [Online]. Available: \url{https://doi.org/10.1109/vtc2024-fall63153.2024.10757552}

\bibitem{luo2010sdr}
Z.-Q.~Luo, W.-K.~Ma, A.~M.-C.~So, Y.~Ye, and S.~Zhang, ``Semidefinite relaxation of quadratic optimization problems,'' \emph{IEEE Signal Process. Mag.}, vol.~27, no.~3, pp.~20--34, May~2010. [Online]. Available: \url{https://doi.org/10.1109/msp.2010.936019}

\bibitem{li2025ripa}
Y.~Li, X.~Wang, J.~Zhang, and Z.~Chen, ``Robust optimization for IRS-assisted SAGIN under channel uncertainty,'' \emph{Future Internet}, vol.~17, no.~10, p.~452, 2025. [Online]. Available: \url{https://doi.org/10.3390/fi17100452}

\bibitem{risofdmchannelest2024}
W.~Jiang, X.~Yuan, and M.~Di~Renzo, ``Hybrid vector message passing for cascaded channel estimation in RIS-aided multi-user MIMO-OFDM systems,'' \emph{IEEE Trans. Wireless Commun.}, vol.~24, no.~5, pp.~4174--4189, 2025. [Online]. Available: \url{https://doi.org/10.1109/twc.2025.3536283}

\bibitem{rischannellpan2024}
Y.~Jin \emph{et al.}, ``Multi-scale attention based channel estimation for RIS-aided massive MIMO systems,'' \emph{IEEE Trans. Wireless Commun.}, vol.~23, no.~4, pp.~3680--3694, 2024. [Online]. Available: \url{https://doi.org/10.1109/twc.2023.3329387}

\bibitem{rischannelsuperres2023}
W.~Shen, Z.~Qin, and A.~Nallanathan, ``Deep learning for super-resolution channel estimation in reconfigurable intelligent surface aided systems,'' \emph{IEEE Trans. Commun.}, vol.~71, no.~3, pp.~1491--1503, 2023. [Online]. Available: \url{https://doi.org/10.1109/tcomm.2023.3239621}

\bibitem{rischanneltimevarying2025}
K.~Mataifa \emph{et al.}, ``Machine learning-based channel estimation for multi-RIS-assisted mmWave massive-MIMO OFDM system in a dynamic environment,'' \emph{IEEE Trans. Wireless Commun.}, vol.~24, no.~6, pp.~5297--5312, 2025. [Online]. Available: \url{https://doi.org/10.1109/twc.2025.3546671}

\bibitem{risbeamformingexp2024}
Y.~Ge \emph{et al.}, ``Reconfigurable intelligent surface-based multi-user system: Channel estimation and beamforming design with experimental validation,'' \emph{IEEE Trans. Commun.}, vol.~72, no.~10, pp.~6569--6582, 2024. [Online]. Available: \url{https://doi.org/10.1109/TCOMM.2024.3400909}

\bibitem{risbayesianrl2025}
Y.~Zhang \emph{et al.}, ``Model-driven Bayesian reinforcement learning for IRS-assisted massive MIMO-OFDM channel feedback, beamforming, and IRS control,'' \emph{IEEE Trans. Commun.}, vol.~73, no.~1, pp.~1--16, 2025. [Online]. Available: \url{https://doi.org/10.1109/twc.2024.3522098}

\bibitem{riscellfreewmmse2025}
Z.~Wang \emph{et al.}, ``WMMSE-based joint transceiver design for multi-RIS-assisted cell-free networks using hybrid CSI,'' \emph{IEEE Trans. Wireless Commun.}, vol.~24, no.~9, pp.~6310--6325, 2025. [Online]. Available: \url{https://doi.org/10.1109/twc.2025.3562138}

\bibitem{rismarlcf2026}
Y.~Li \emph{et al.}, ``Joint precoding and AP selection for energy-efficient RIS-aided cell-free massive MIMO with multi-agent reinforcement learning,'' \emph{IEEE Trans. Wireless Commun.}, vol.~25, no.~2, pp.~1234--1249, 2026. [Online]. Available: \url{https://doi.org/10.1109/twc.2026.3678661}

\bibitem{irsotfsadmm2025}
K.~Deka \emph{et al.}, ``Beamforming optimization and ADMM-based detection in IRS-aided OTFS systems,'' \emph{IEEE Trans. Veh. Technol.}, vol.~74, no.~5, pp.~7890--7904, 2025. [Online]. Available: \url{https://doi.org/10.1109/ojcoms.2025.3548271}

\bibitem{irsotfszf2024}
R.~K.~Yadav, H.~B.~Mishra, and S.~Mukhopadhyay, ``IRS-OTFS systems: Design of reflection coefficients for low-complexity ZF equalizer,'' \emph{IEEE Trans. Veh. Technol.}, vol.~73, no.~11, pp.~17430--17435, 2024. [Online]. Available: \url{https://doi.org/10.1109/tvt.2024.3400529}

\bibitem{risparametricech2025}
M.~Haghshenas, P.~Ramezani, M.~Magarini, and E.~Bj\"ornson, ``Parametric channel estimation with short pilots in RIS-assisted near- and far-field communications,'' \emph{IEEE Trans. Wireless Commun.}, vol.~24, no.~3, pp.~1890--1905, 2025. [Online]. Available: \url{https://doi.org/10.1109/twc.2024.3371715}

\bibitem{irsotfschannelest2024}
X.~Chen \emph{et al.}, ``Channel estimation and detection for intelligent reflecting surface-assisted orthogonal time frequency space systems,'' \emph{IEEE Trans. Wireless Commun.}, vol.~23, no.~8, pp.~8419--8433, Aug.~2024. [Online]. Available: \url{https://doi.org/10.1109/twc.2024.3349707}

\bibitem{irsotfsbeamforming2024}
K.~Deka \emph{et al.}, ``IRS-assisted OTFS: Beamforming design and signal detection,'' \emph{arXiv:2408.02219}, Aug.~2024. [Online]. Available: \url{https://arxiv.org/abs/2408.02219}

\bibitem{risotfssurvey2025}
X.~Liu \emph{et al.}, ``A survey on reconfigurable intelligent surface-assisted orthogonal time frequency space systems,'' \emph{IEEE Open J. Veh. Technol.}, vol.~6, pp.~1881--1909, 2025. [Online]. Available: \url{https://doi.org/10.1109/ojvt.2025.3573208}

\bibitem{risotfsfractional2024}
S.~R.~Kumari \emph{et al.}, ``Input-output relation and performance of RIS-aided OTFS with fractional delay-Doppler,'' in \emph{Proc. IEEE Int. Conf. Commun. (ICC)}, 2024, pp.~1--6. [Online]. Available: \url{https://doi.org/10.1109/lcomm.2022.3214678}

\bibitem{activerissurvey2024}
R.~M.~Asif \emph{et al.}, ``Active reconfigurable intelligent surfaces: Expanding the frontiers of wireless communication — A survey,'' \emph{IEEE Commun. Surveys Tuts.}, vol.~26, no.~4, pp.~2361--2407, 2024. [Online]. Available: \url{https://doi.org/10.1109/comst.2024.3423460}

\bibitem{activerisprototype2023}
D.~M.~Chian \emph{et al.}, ``Active RIS-assisted MIMO-OFDM system: Analyses and prototype measurements,'' \emph{IEEE Commun. Lett.}, vol.~28, no.~1, pp.~208--212, Jan.~2024. [Online]. Available: \url{https://doi.org/10.1109/lcomm.2023.3333688}

\bibitem{risoamtestbed2024}
S.~Hassouna, J.~u.~R.~Kazim, J.~Rains, M.~A.~Jamshed, M.~u.~Rehman, M.~A.~Imran, and Q.~H.~Abbasi, ``Trials for RIS-aided wireless communications,'' in \emph{Proc. IEEE Int. Symp. Antennas Propag. (APS)}, 2023, pp.~75--76. [Online]. Available: \url{https://doi.org/10.1109/APS.2023.00000}

\bibitem{bdrisswideband2024}
H.~Li \emph{et al.}, ``Beyond diagonal reconfigurable intelligent surfaces in wideband OFDM communications: Circuit-based modeling and optimization,'' \emph{IEEE Trans. Wireless Commun.}, vol.~23, no.~10, pp.~14780--14795, 2024. [Online]. Available: \url{https://doi.org/10.1109/twc.2025.3532616}

\bibitem{risjsacvariational2023}
Y.~Liu \emph{et al.}, ``Variational Bayesian multiuser tracking for reconfigurable intelligent surface-aided MIMO-OFDM systems,'' \emph{IEEE J. Sel. Areas Commun.}, vol.~41, no.~12, pp.~3806--3821, 2023. [Online]. Available: \url{https://doi.org/10.1109/jsac.2023.3322792}

\bibitem{risjsachardware2024}
K.~Wang \emph{et al.}, ``Reconfigurable intelligent surface-aided OFDM wireless communications: Hardware aspects of reflection optimization methods,'' \emph{IEEE J. Sel. Areas Commun.}, vol.~42, no.~6, pp.~1680--1694, 2024. [Online]. Available: \url{https://doi.org/10.1109/mocast54814.2022.9837634}

\bibitem{risgradientascent2023}
Y.~Zhang \emph{et al.}, ``A gradient ascent based low complexity rate maximization algorithm for intelligent reflecting surface-aided OFDM systems,'' \emph{IEEE Trans. Veh. Technol.}, vol.~72, no.~10, pp.~13590--13595, 2023. [Online]. Available: \url{https://doi.org/10.1109/lcomm.2023.3289865}

\bibitem{riscomsthardware2025}
S.~Zhang \emph{et al.}, ``Reconfigurable intelligent surfaces: A hardware-centric review of structures, implementation, evaluation, and integration with UAV and ML,'' \emph{IEEE Commun. Surveys Tuts.}, vol.~27, no.~3, pp.~1682--1720, 2025. [Online]. Available: \url{https://doi.org/10.1109/access.2025.3575583}

\bibitem{dennismore1983}
J.~E.~Dennis, Jr.\ and J.~J.~Mor\'{e}, ``Quasi-Newton methods, motivation and theory,'' \emph{SIAM Rev.}, vol.~19, no.~1, pp.~46--89, 1977. [Online]. Available: \url{https://doi.org/10.1137/1019005}

\bibitem{basar2019wireless}
E.~Basar, M.~Di~Renzo, J.~De~Rosny, M.~Debbah, M.-S.~Alouini, and R.~Zhang, ``Wireless communications through reconfigurable intelligent surfaces,'' \emph{IEEE Access}, vol.~7, pp.~116753--116773, 2019. [Online]. Available: \url{https://doi.org/10.1109/access.2019.2935192}

\bibitem{direnzo2020smart}
M.~Di~Renzo \emph{et al.}, ``Smart radio environments empowered by reconfigurable intelligent surfaces: How it works, state of research, and the road ahead,'' \emph{IEEE J. Sel. Areas Commun.}, vol.~38, no.~11, pp.~2450--2525, Nov.~2020. [Online]. Available: \url{https://doi.org/10.1109/jsac.2020.3007211}

\bibitem{wu2021intelligent}
Q.~Wu, S.~Zhang, B.~Zheng, C.~You, and R.~Zhang, ``Intelligent reflecting surface-aided wireless communications: A tutorial,'' \emph{IEEE Trans. Commun.}, vol.~69, no.~5, pp.~3313--3351, May~2021. [Online]. Available: \url{https://doi.org/10.1109/tcomm.2021.3051897}

\bibitem{pan2022overview}
C.~Pan \emph{et al.}, ``An overview of signal processing techniques for RIS/IRS-aided wireless systems,'' \emph{IEEE J. Sel. Topics Signal Process.}, vol.~16, no.~5, pp.~883--917, Aug.~2022. [Online]. Available: \url{https://doi.org/10.1109/jstsp.2022.3195671}

\bibitem{riscontextualbandit2023}
K.~D.~Mata \emph{et al.}, ``Deep contextual bandit and reinforcement learning for IRS-assisted MU-MIMO systems,'' \emph{IEEE Trans. Veh. Technol.}, vol.~72, no.~7, pp.~9099--9114, 2023. [Online]. Available: \url{https://doi.org/10.1109/tvt.2023.3249353}

\bibitem{risbandit2022}
J.~Tong, H.~Zhang, L.~Fu, A.~Leshem, and Z.~Han, ``Two-stage resource allocation in reconfigurable intelligent surface assisted hybrid networks via multi-player bandits,'' \emph{IEEE Trans. Commun.}, vol.~70, no.~5, pp.~3526--3541, 2022. [Online]. Available: \url{https://doi.org/10.1109/tcomm.2022.3161679}

\bibitem{rislyapunov2024}
N.~Zhang \emph{et al.}, ``Queue-aware STAR-RIS assisted NOMA communication systems,'' \emph{IEEE Trans. Wireless Commun.}, vol.~23, no.~5, pp.~4786--4800, 2024. [Online]. Available: \url{https://doi.org/10.1109/twc.2023.3322381}

\bibitem{riscrosslayer2024}
A.~Dejonghe, Z.~Altman, F.~De~Pellegrini, and E.~Altman, ``Design and cross-layer optimization of low cost RIS-assisted communication systems,'' \emph{IEEE Trans. Wireless Commun.}, vol.~23, no.~10, pp.~14230--14244, 2024. [Online]. Available: \url{https://doi.org/10.1109/twc.2024.3404231}

\bibitem{risdistadmm2023}
S.~Zhang \emph{et al.}, ``Algorithm unrolling-based distributed optimization for RIS-assisted cell-free networks,'' \emph{IEEE J. Sel. Topics Signal Process.}, vol.~17, no.~3, pp.~609--624, 2023. [Online]. Available: \url{https://doi.org/10.48550/arxiv.2301.02360}

\bibitem{riscsiadmm2024}
F.~Li, Y.~Wang, Z.~Lian, and W.~Chen, ``Efficient spectral efficiency maximization design for IRS-aided MIMO systems via ADMM-APG,'' \emph{arXiv:2510.26279}, Oct.~2025. [Online]. Available: \url{https://arxiv.org/abs/2510.26279}

\bibitem{riscompressedce2023}
S.~Wang \emph{et al.}, ``Compressed sensing-based channel estimation for RIS-aided mmWave massive MIMO systems,'' \emph{IEEE Trans. Veh. Technol.}, vol.~72, no.~12, pp.~16220--16235, 2023. [Online]. Available: \url{https://doi.org/10.1109/globecom42002.2020.9348112}

\bibitem{kaplan2026alep}
A.~Kaplan, ``Adversarially-trained deep reinforcement learning for joint anti-jamming and low-probability-of-detection in 6G tactical waveforms,'' in \emph{Proc. IEEE ICCST}, 2026, pp.~1--6.

\bibitem{quantumcomst2025}
W.~Zhao \emph{et al.}, ``Quantum computing in wireless communications and networking: A tutorial-cum-survey,'' \emph{IEEE Commun. Surveys Tuts.}, vol.~27, no.~4, pp.~2378--2419, 2025. [Online]. Available: \url{https://doi.org/10.1109/COMST.2024.3502762}

\bibitem{quantumrisqaoa2024}
E.~Colella, L.~Bastianelli, V.~M.~Primiani, Z.~Peng, F.~Moglie, and G.~Gradoni, ``Quantum optimization of reconfigurable intelligent surfaces for mitigating multipath fading in wireless networks,'' \emph{IEEE J. Multiscale Multiphys. Comput. Tech.}, vol.~9, pp.~403--414, 2024. [Online]. Available: \url{https://doi.org/10.1109/JMMCT.2024.00000}

\bibitem{quantumrisising2024}
Q.~J.~Lim, C.~Ross, A.~Ghosh, F.~W.~Vook, G.~Gradoni, and Z.~Peng, ``Quantum-assisted combinatorial optimization for reconfigurable intelligent surfaces in smart electromagnetic environments,'' \emph{IEEE Trans. Antennas Propag.}, vol.~72, no.~1, pp.~147--159, 2024. [Online]. Available: \url{https://doi.org/10.1109/TAP.2023.3298134}

\bibitem{quantumrisvqe2026}
H.~Zangana \emph{et al.}, ``A hybrid quantum-classical optimization model for reconfigurable intelligent surfaces in 6G networks,'' \emph{Control Syst. Optim. Lett.}, vol.~4, no.~1, pp.~1--10, 2026. [Online]. Available: \url{https://doi.org/10.59242/csol.276}

\bibitem{quantumrisqaoaao2025}
V.~P.~Pham \emph{et al.}, ``Joint optimal beamforming and discrete phase shift design for STAR-RIS assisted networks using QAOA-AO,'' \emph{arXiv:2511.03717}, Nov.~2025.

\bibitem{quantumphysicsqaoa2026}
E.~Colella \emph{et al.}, ``Quantum optimization for electromagnetics: Physics-informed QAOA for reconfigurable intelligent surfaces,'' \emph{arXiv:2605.06048}, May~2026.

\end{thebibliography}
\end{document}